\documentclass{article} % For LaTeX2e
\usepackage{iclr2023_conference,times}

\usepackage{amsmath,amsfonts,bm}

\def\eqref#1{equation~\ref{#1}}
\def\1{\bm{1}}

\DeclareMathAlphabet{\mathsfit}{\encodingdefault}{\sfdefault}{m}{sl}
\SetMathAlphabet{\mathsfit}{bold}{\encodingdefault}{\sfdefault}{bx}{n}

\DeclareMathOperator*{\argmax}{arg\,max}

\usepackage{hyperref}
\usepackage{url}
\usepackage{graphicx}
\usepackage{tabularx}
\usepackage{subfigure}
\usepackage[ruled, lined, linesnumbered, commentsnumbered, longend]{algorithm2e}
\usepackage{multicol}
\usepackage{multirow}
\usepackage{amsmath}
\usepackage{xcolor}
\usepackage{color}
\newcommand{\xgy}[1]{{\color{black} #1}}

\title{Pushing the Limits of Few-shot Anomaly Detection in Industry Vision: Graphcore}
\author{Guoyang Xie$^{1,2,}$\thanks{Contributed Equally, $^{\dag}$Corresponding Authors.}, Jinbao Wang$^{1,*,\dag}$, Jiaqi Liu$^{1,*}$, Feng Zheng$^{1,\dag}$, Yaochu Jin$^{2,3}$ \\
$^{1}$Research Institute of Trustworthy Autonomous Systems, \\ Southern University of Science and Technology, Shenzhen 518055, China \\ $^{2}$NICE Group, University of Surrey, Guildford, GU2 7YX, United Kingdom \\$^{3}$NICE Group, Bielefeld University, Bielefeld 33619, Germany \\
\texttt{guoyang.xie@surrey.ac.uk}, \texttt{linkingring@163.com} \\ 
\texttt{liujq32021@mail.sustech.edu.cn}, \texttt{zhengf@sustech.edu.cn} \\ \texttt{yaochu.jin@uni-bielefeld.de}
}
\iclrfinalcopy % Uncomment for camera-ready version, but NOT for submission.
\begin{document}

\maketitle

\begin{abstract}

% 瑕疵检测邻域里面，结构特征在以memory bank为代表的方法中起到了至关重要的作用，同时特征的特性也限制了瑕疵检测的性能。我们发现旋转不变的特征对瑕疵起到了重要作用。为此，本文探究了基于图表征的在瑕疵检测任务的应用，提出了一种新的视觉等距不变特征，这一特性具有旋转不变性，能够提升特征鲁棒性，进一步显著地降低M中存储的冗余特征数量。本文提出了基于VIIF设计的GraphCore模型，其能够有效地实现无监督的小样本瑕疵检测，显著地提升了模型检测性能。实验结果和可视化表明我们提出的方法是有效的，在三个数据集上取得了最好的准确性。
% 然而现在这些方法没有考虑特征与工业品数据的关系，存在瑕疵特征判别与特征之间不匹配的问题，极大地限制了性能。
% 
% here, can we provide a problem of existing methods????

In the area of few-shot anomaly detection (FSAD), efficient visual feature plays an essential role in the memory bank $\mathcal{M}$-based methods. However, these methods do not account for the relationship between the visual feature and its rotated visual feature, drastically limiting the anomaly detection performance.
To push the limits, we reveal that rotation-invariant feature property has a significant impact on industrial-based FSAD. 
Specifically, we \xgy{utilize} graph representation in FSAD and provide a novel visual isometric invariant feature (VIIF) as an anomaly measurement feature. 
As a result, VIIF can robustly improve the anomaly discriminating ability and can further reduce the size of redundant features stored in $\mathcal{M}$ by a large amount. Besides, we provide a novel model GraphCore via VIIFs that can fast implement unsupervised FSAD training and improve the performance of anomaly detection. A comprehensive evaluation is provided for comparing GraphCore and other SOTA anomaly detection models under our proposed few-shot anomaly detection setting, which shows GraphCore can increase average AUC by 5.8\%, 4.1\%, 3.4\%, and 1.6\% on MVTec AD and by 25.5\%, 22.0\%, 16.9\%, and 14.1\% on MPDD for 1, 2, 4, and 8-shot cases, respectively.

\end{abstract}

\section{Introduction}

% 随着人工智能深度视觉检测技术的快速发展，面向工业品表面的瑕疵（包括划痕、污点、缺失等）检测得到了空前的关注。在实际生产环节中，一个重要的问题是如何实现转产场景下工业品模型的快速训练同时保证瑕疵的精准检测。

% 目前在瑕疵检测领域，存在的现状如下：
% 在检测准确性方面，当前主流的工作采用大量的训练数据作为输入去训练模型，如图1(a)，但是会导致数据收集困难，如果采用全监督的方式，还会存在标注成本高。为此，一些以牺牲准确性为代价基于小样本学习的方法被提出，如xxx，采用元学习的方式，如图1(b)，但是设置复杂，无法灵活地迁移到转产后的新产品，检测性能无法保证。
% 在训练速度方面，如果采用大量的数据进行训练，势必导致转产后对新产品的训练速度变慢，无法在真实的生产环节。目前通用的异常检测需要收集大量数据，而在小样本学习中，尽管采用元学习的方式，如图1(b)，其仍需要训练大量前期收集的数据。
% 如图1(c)为本文提出的小样本瑕疵检测新方法，只采用少量数据实现转产产品的快速训练和可竞争的瑕疵检测准确性能。我们方法的优势是，一反面，由于只采用少量数据，可以快速训练和测试时的推理速度，二是因为直接对新产品样本进行训练，不存在瑕疵需要适应和迁移的问题。

%% 讨论视觉等距不变特征对瑕疵检测的必要性
% 基于一个事实，即工业品生产出来具有高稳定性（外形颜色不发生明显变形），从而拍摄的图片不具有自然图像的多样性，更多的是存在拍摄角度的问题。我们声称面对工业品的瑕疵检测仅需要少量数据即可实现与大量数据一致的效果，少量的图像数据即可包含大量数据已有的信息。为此，我们在本文中提出的小样本学习方法的核心思想是采用视觉等距不变的特征作为瑕疵度量特征。在以Memory Bank（M）为瑕疵检测范式的方法中（如PatchCore），其采用ImageNet/ResNet作为特征提取器，但是由于卷积操作获得的特征不具有旋转不变性（证明卷积不具有旋转不变性的文章），所以在M中存储了大量的冗余特征，如同一个小结构的多个旋转特性。就会导致需要大量的训练数据才能保证测试集的高准确性。基于此，为了消除大量的冗余特性，我们提出的视觉等距不变特征，一方面获得更鲁棒的视觉特征，另一方面可以显著降低M大小，加快检测速度，解决基于PatchCore方法M太大的问题。

% 从实验结果证明视觉等距不变特征的有效性，显著提升了xxx数据集的性能。

With the rapid development of deep vision detection technology in artificial intelligence, detecting anomalies/defects on the surface of industrial products has received unprecedented attention. Changeover in manufacturing refers to converting a line or machine from processing one product to another. Since the equipment has not been completely fine-tuned after the start of the production line, changeover frequently results in unsatisfactory anomaly detection (AD) performance.

How to achieve rapid training of industrial product models in the changeover scenario while assuring accurate anomaly detection is a critical issue in the actual production process. The current state of AD in the industry is as follows:
% If full supervision is adopted, there will also be substantial labeling expenses. 
(1) In terms of detection accuracy, the performance of state-of-the-art (SOTA) AD models degrades dramatically during the changeover. Current mainstream work utilizes a considerable amount of training data as input to train the model, as shown in Fig.~1(a). However, this will make data collecting challenging, even for unsupervised learning. As a result, many approaches based on few-shot learning at the price of accuracy have been proposed. For instance, \cite{huang2022registration} employ meta-learning, as shown in Fig.~1(b). While due to complicated settings, it is impossible to migrate to the new product during the changeover flexibly, and the detection accuracy cannot be guaranteed.
(2) In terms of training speed, when a large amount of data is utilized for training, the training progress for new goods is slowed in the actual production line. As is well-known, vanilla unsupervised AD requires to collect a large amount of information. Even though meta-learning works in few-shot learning, as shown in Fig.~1(b), it is still necessary to train a massive portion of previously collected data.

We state that AD of industrial products requires just a small quantity of data to achieve performance comparable to a large amount of data, i.e., a small quantity of image data can contain sufficient information to represent a large number of data. Due to the fact that industrial products are manufactured with high stability (no evident distortion of shape and color cast), the taken images lack the diversity of natural images, and there is a problem with the shooting angle or rotation. Therefore, it is essential to extract rotation-invariant structural features. As graph neural networks (GNNs) are capable of robustly extracting non-serialized structural features (\cite{han2022vision}, \cite{bruna2013spectral}, \cite{hamilton2017inductive}, \cite{xu2018powerful}), and they integrate global information better and faster~\cite{wang2020haar,li2020fast}. They are more suited than convolution neural networks (CNNs) to handle the problem of extracting rotation-invariant features. For this reason, the \textit{core idea} of the proposed \textit{GraphCore} method in this paper is to use the visual isometric invariant features (VIIFs) as the anomaly measurement features. In the method using memory bank ($\mathcal{M}$) as the AD paradigm,  PatchCore ~(\cite{roth2022towards}) uses ResNet~(\cite{He2016IdentityMI}) as the feature extractor. However, since their features obtained by CNNs do not have rotation invariance (\cite{dieleman2016exploiting}), a large number of redundant features are stored in $\mathcal{M}$. Note that these redundant features maybe come from multiple rotation features of the same patch structure. It will hence require a huge quantity of training data to ensure the high accuracy of the test set. To avoid these redundant features, we propose VIIFs, which not only produce more robust visual features but also dramatically lower the size of $\mathcal{M}$ and accelerate detection.

Based on the previous considerations, the goal of our work is to handle the cold start of the production line during the changeover. As shown in Fig.~1(c), a new FSAD method, called \textit{GraphCore}, is developed that employs a small number of normal samples to accomplish fast training and competitive AD accuracy performance of the new product. On the one hand, by utilizing a small amount of data, we would rapidly train and accelerate the speed of anomaly inference. On the other hand, because we directly train new product samples, adaptation and migration of anomalies from the old product to the new product do not occur.

\begin{figure}
    \centering
    \includegraphics[width=0.85\linewidth]{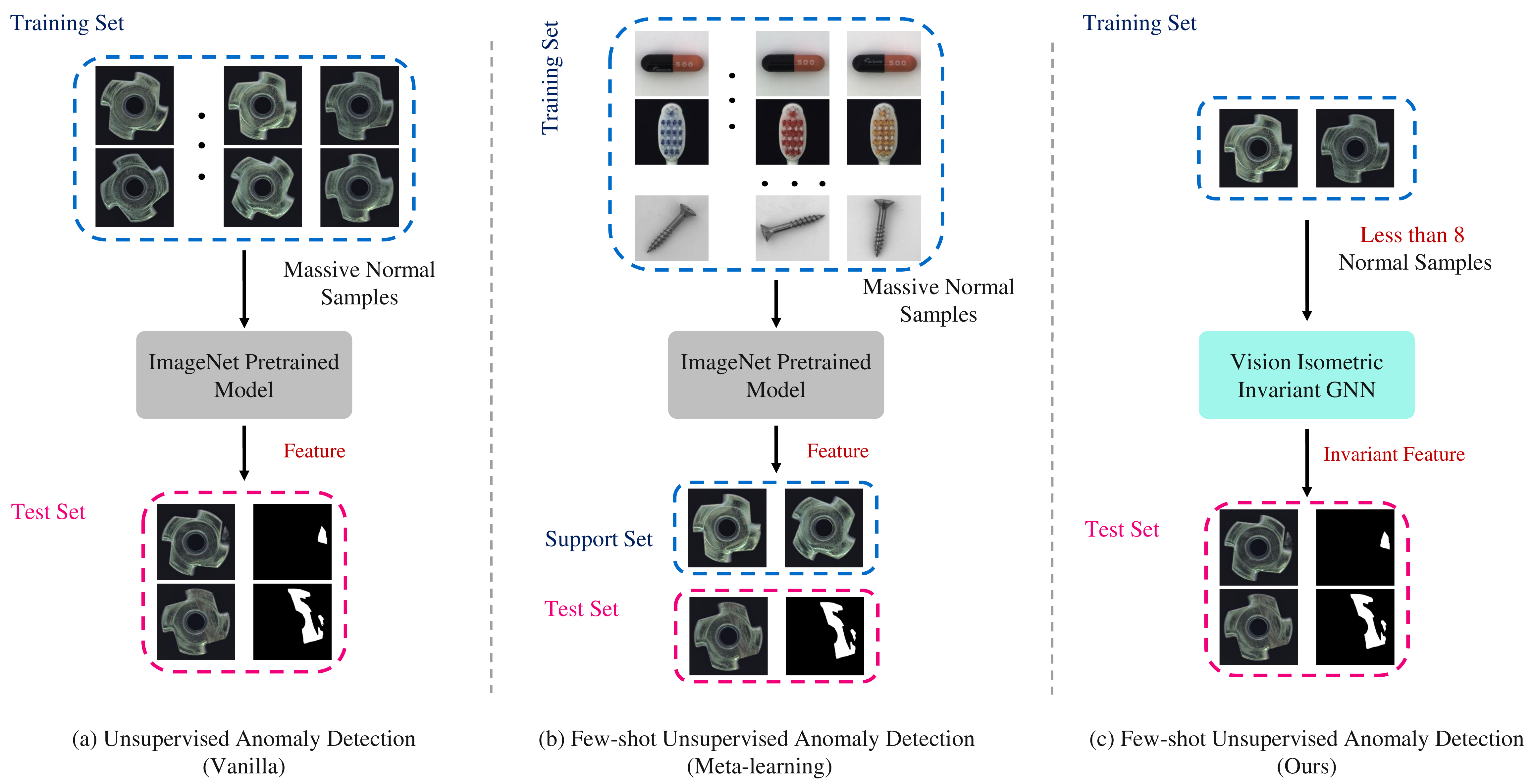}
    \caption{Different from (a) vanilla unsupervised AD and (b) few-shot unsupervised AD in meta learning. As input training samples, our setting (c) only utilizes a small number of normal samples. For our setting (c), there is no requirement to aggregate training categories in advance. The proposed model, vision isometric invariant GNN, can fast obtain the invariant feature within a few normal samples, and its accuracy outperforms models trained in a meta-learning context.}
    \label{fig:ad_paradigm}
\end{figure}

\textbf{Contributions.} In summary, the main contributions of this work are as follows:
\begin{itemize}
    \item We present a feature-augmented method for FSAD in order to investigate the property of visual features generated by CNNs.
    \item We propose a novel anomaly detection model, GraphCore, to add a new VIIF into the memory bank-based AD paradigm, which can drastically reduce the quantity of redundant visual features.
    %\item Based on VIIFs, we propose a new method, GraphCore, to achieve few-shot AD. 
    \item The experimental results show that the proposed VIIFs are effective and can significantly enhance the FSAD performance on MVTec AD and MPDD.
\end{itemize}

\textbf{Related Work.} Few-shot anomaly detection (FSAD) is an attractive research topic. However, there are only a few papers devoted to the industrial image FSAD. Some works (\cite{liznerski2020explainable,pang2021explainable,ding2022catching}) experiment with few-shot abnormal images in the test set, which contradicts our assumptions that no abnormal images existed. While others (\cite{wu2021learning,huang2022registration}) conduct experiments in a meta-learning setting. This configuration has the disadvantage of requiring a high number of base class images and being incapable of addressing the shortage of data under cold-start conditions in industrial applications. PatchCore (\cite{roth2022towards}), SPADE (\cite{cohen2020sub}), and PaDiM (\cite{defard2021padim}) investigated AD performance on MVTec AD in a few-shot setting. However, these approaches are not intended for changeover-based few-shot settings. Thus their performance cannot satisfy the requirements of manufacturing changeover. In this research, we propose a feature augmentation method for FSAD that can rapidly finish the training of anomaly detection models with a small quantity of data and meet manufacturing changeover requirements.

%CN: Jinbao TODO: 最后两句话需要重新写，英文太过中式了，这是佳奇写的

%CN:\cite{pang2021explainable,ding2022catching,liznerski2020explainable}借助了测试集中的异常图像，这并不符合我们的设定。\cite{wu2021learning,huang2022registration}在meta-learning的设定下进行了实验，这一设定的最大问题在于需要借助大量的基类图像，无法满足工业场景中冷启动的问题。\cite{rudolph2021same,sheynin2021hierarchical}中few-shot的设定与实验与我们较为相似，但最主要的问题在于模型训练耗时过久，且性能较差，无法满足工业转场过程中快捷高效的要求。pathcore和padim也在few-shot场景下进行了实验，但没有针对性设置，因而性能难以满足转场需求。
% \cite{ruff2019deep}Deep semi-supervised anomaly detection 非图像异常，不说明
% A hierarchical transformation-discriminating generative model for few-shot anomaly detection \cite{sheynin2021hierarchical}
% Same same but differnet: Semi-supervised defect detection with normalizing flows.\cite{rudolph2021same}
% \cite{pang2021explainable}Explainable Deep few-shot Anomaly Detection with Deviation Networks
% \cite{ding2022catching}Catching Both Gray and Black Swans: Open-set Supervised Anomaly Detection
% \cite{wu2021learning,huang2022registration} evaluate, but which paper is not suitable for changeover in manufacturing. 

\section{Approach}

\textbf{Problem Setting.}\label{sec:challenge}
Fig.~\ref{fig:ad_paradigm}(c) outlines the formal definition of the problem setting for the proposed FSAD. Given a training set of only $n$ normal samples during training, where $n \leq  8$, from a specific category. At test time, given a normal or abnormal sample from a target category, the anomaly detection model should predict whether or not the image is anomalous and localize the anomaly region if the prediction result is anomalous. 

\textbf{Challenges.} For the FSAD proposed in Fig.~\ref{fig:ad_paradigm}(c), we attempt to detect anomalies in the test sample using only a small number of normal images as the training dataset. The key challenges consist of: (1) Each category's training dataset contains only normal samples, i.e., no annotations at the image or pixel level. (2) There are few normal samples of the training set available. In our proposed setting, there are fewer than 8 training samples.

\begin{figure}[ht]
    \centering
    \includegraphics[width=0.7\linewidth]{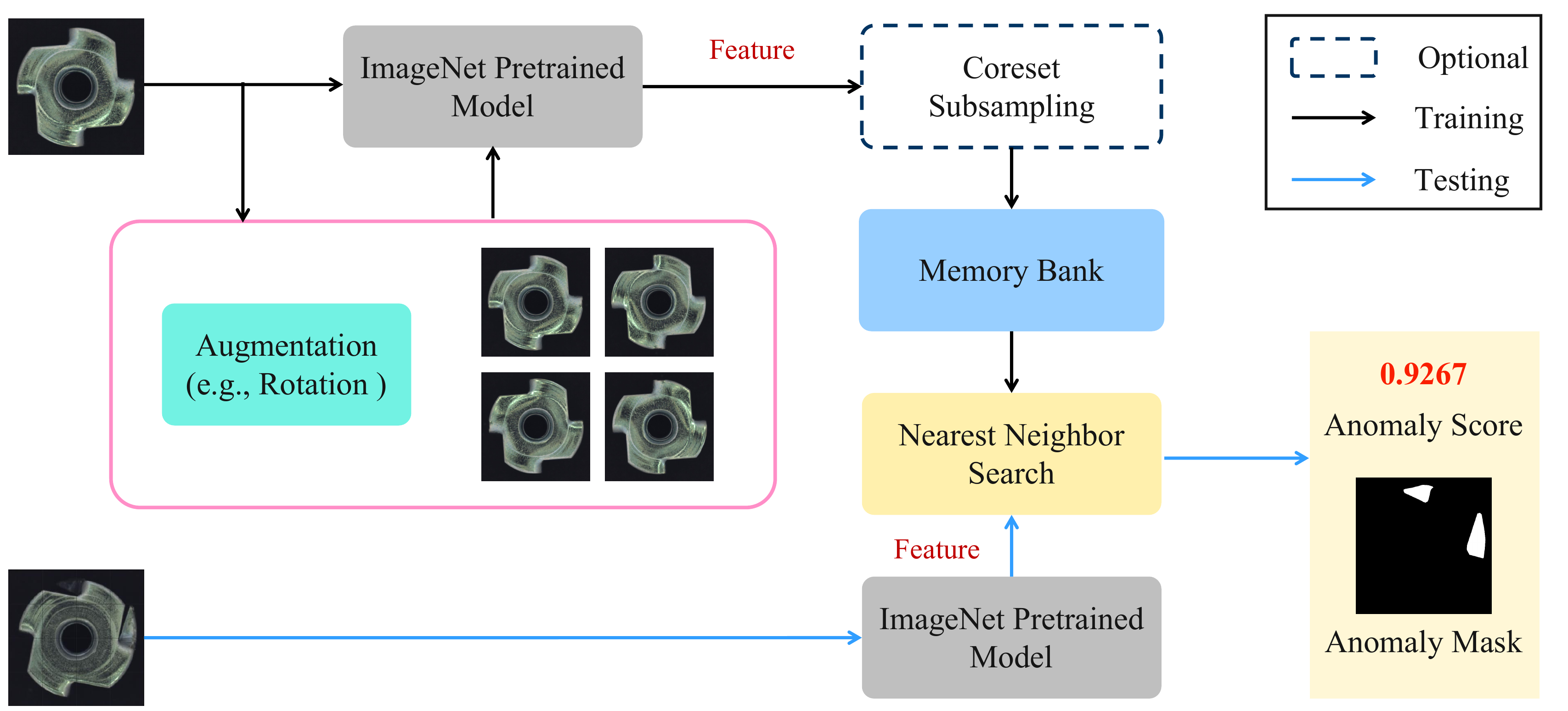}
    \caption{Augmentation+PatchCore Architecture.}
    \label{fig:rotation_augmentation}
\end{figure}

%CN:先搞几个旋转和平移起来看的很像的图片。(图4中的metal nut 和图6中的钉子)，以他们为例子，看着这些图片，motivation就油然而生了。因为手上的数据非常少, 所以看到这种图片，其实是可以通过简单的rotation augmentation, 就可以复制到这些feature. 
\textbf{Motivation.} In the realistic industrial image dataset (\cite{Bergmann2019MVTecA,jezek2021deep}), the images under certain categories are extremely similar. Most of them can be converted to one another with simple data augmentation, such as the meta nut (Fig.~\ref{fig:rotation_augmentation}) and \xgy{the screw (Fig.~\ref{fig:unified_view})}. For instance, rotation augmentation can effectively provide a new screw dataset. Consequently, when faced with the challenges stated in Section~\ref{sec:challenge}, our natural inclination is to acquire additional data through data augmentation. Then, the feature memory bank (Fig.~\ref{fig:gnn_arch}) can store more useful features.

\subsection{Augmentation+PatchCore}\label{sec:aug_patchcore} 
To validate our insight, we have adapted PatchCore~(\cite{roth2022towards}) to our model. We denote augmentation (rotation) with PatchCore as Aug.(R). The architecture is depicted in detail in Fig.~\ref{fig:rotation_augmentation}. Before extracting features from the ImageNet pre-trained model, we augment the data (e.g., by rotating the data).

\begin{algorithm}[th]
\small
\caption{Aug.(R) memory bank}\label{alg:aug}
    \SetKwInOut{KwIn}{Input}
    \SetKwInOut{KwOut}{Output}
    \KwIn{ImageNet pre-trained $\phi$, all normal samples $\mathcal{X}_{N}$, data augmentation operator $\alpha$, patch feature extractor $\mathcal{P}$, memory size target $l$, random linear projection $\psi$.}
    \KwOut{Patch-level augmented memory bank $\mathcal{M}$.}
    $\mathcal{M} \leftarrow \left\{ \right\}$\;
    \For{$x_{i} \in \mathcal{X_{N}}$}
    {
        $x^{g}_{i} \leftarrow \alpha(x_{i})$\;  
        $\mathcal{M} \leftarrow \mathcal{P}(\phi(x_{i}))$\;
        $\mathcal{M} \leftarrow \mathcal{P}(\phi(x^{g}_{i}))$\;
    }
    $\mathcal{M}_{C} \leftarrow \left\{ \right\}$ \textcolor{gray}{//Apply coreset sampling for memory bank}\\
    \For{$i \in \left [0,\cdots, l-1  \right ]$}
    {
        $m_{i} \leftarrow \underset{m \in \mathcal{M} - \mathcal{M_{C}}}{\argmax} \underset{n \in \mathcal{M_{C}}} {\min} \left\| \psi(m) - \psi(n) \right\|_{2}$\;
        $\mathcal{M_{C}} \leftarrow \mathcal{M_{C}} \cup \left\{m_{i} \right\}$\;
    }
    $\mathcal{M} \leftarrow \mathcal{M_{C}}$.
\end{algorithm}

In the training phase, the aim of the training phase is to build up a memory bank, which stores the neighborhood-aware features from all normal samples. At test time, the test image is predicted as anomalies if at least one patch is anomalous, and pixel-level anomaly segmentation is computed via the score of each patch feature. The feature memory construction method is shown in Algorithm~\ref{alg:aug}. 
% Namely, $\mathcal{P}$ denotes the patch feature extraction from each normal sample $x_{i} \in \mathcal{X_{N}}$. 
We default set ResNet18 (\cite{He2016IdentityMI}) as the feature extraction model. Conceptually, coreset sampling (\cite{Sener2018ActiveLF}) for memory bank aims to balance the size of the memory bank with the performance of anomaly detection. And the size of the memory bank has a considerable impact on the inference speed. In Section~\ref{sec:sampling}, we discuss the effect of the sampling rate in detail. 

In testing phase, with the normal patches feature bank $\mathcal{M}$, the image-level anomaly score $s$ for the test image $x^{test}$ is computed by the maximum score $s^{*}$ between the test image's patch feature $\mathcal{P}(x^{test})$ and its respective nearest neighbour $m^{*}$ in $\mathcal{M}$.

% Namely, Aug.(R) denotes the rotation augmentation+PatchCore. 
From Table~\ref{tab:total_metric} and Table~\ref{tab:mvtec-2shot}, we can easily observe that the performance of Aug.(R) greatly outperforms the SOTA models under the proposed few-shot setting.

\begin{figure}[b]
    \centering
    \includegraphics[width=0.55\linewidth]{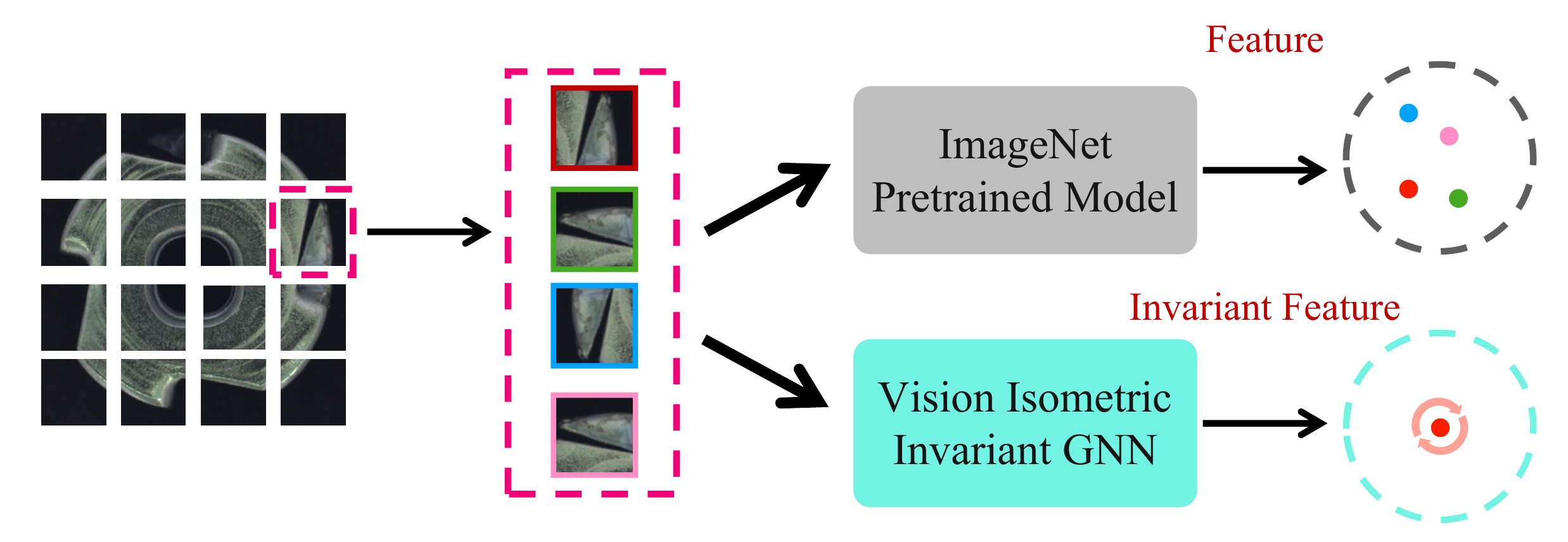}
    \caption{Convolution feature VS vision isometric invariant feature.}
    \label{fig:gnn_motivation}
\end{figure}

\subsection{Vision Isometric Invariant Feature}
%CN:在上面那一段，我们经验性地证明了数据增强+PatchCore是有效的。那我们追究本质，为什么这种方法会如此有效，能不能从一个更加通用的方法来考虑呢？

% In Section~\ref{sec:aug_patchcore}, we heuristically prove that Augmentation+PatchCore outperform the SOTA models under the proposed few-shot anomaly detection setting. Essentially, data augmentation method directly adds normal samples' features into the memory bank. In other words, Augmentation+PatchCore enhance the possibility to find a subset feature such that the anomaly score of test image can be more accurately calculated via Equ.~(\ref{eq:find_neighbour}) and Equ.~(\ref{eq:anomaly_score}). So we doubt that is it possible to obtain the invariant representational features from a few normal samples and add them into the feature memory bank? To this end, we propose a novel feature extraction model--vision isometric invariant graph neural network (VIIG), which is shown in Fig.~\ref{fig:gnn_motivation}. Inspired by Section~\ref{sec:challenge}, the proposed model aims to extract visual isometric invariant feature (VIIF) from each patch of the normal sample. As we mentioned before, most of industrial visual anomaly detection dataset is able to transformed to each other by rotation, translation and flipping. Thus, the isomorphism of GNN perfectly fit in the industrial visual anomaly detection.

In Section~\ref{sec:aug_patchcore}, we heuristically demonstrate that Augmentation+PatchCore outperforms SOTA models in the few-shot anomaly detection context proposed. Essentially, the data augmentation method immediately incorporates the features of normal samples into the memory bank. In other words, Augmentation+PatchCore improves the probability of locating a subset feature, allowing the anomaly score of the test image to be calculated with greater precision. Therefore, we question whether it is possible to extract the invariant representational features from a small number of normal samples and add them to the feature memory bank. As demonstrated in Fig.~\ref{fig:gnn_motivation}, we propose a new model for feature extraction: vision isometric invariant graph neural network (VIIG). The proposed model is motivated by  Section~\ref{sec:challenge} and tries to extract the visual isometric invariant feature (VIIF) from each patch of the normal sample. As previously stated, the majority of industrial visual anomaly detection datasets are transformable via rotation, translation, and flipping. Thus, the isomorphism of GNN suited industrial visual anomaly detection excellently.

% \begin{figure}[htb]
% \centering
% \subfigure[Aug.(R)]{
%     \label{fig:architurcture-augr}
%     \includegraphics[width=0.49\textwidth]{iclr2023/fig/rotation_augmentation.pdf}}
%         % \hspace{-2mm}
% \subfigure[GraphCore]{
%     \label{fig:architurcture-graphcore}
%     \includegraphics[width=0.49\textwidth]{iclr2023/fig/gnn_pipeline.pdf}}
%         % \hspace{-2mm}
% \caption{Rotation augmentation and vision isometric invariant GNN for few-shot anomaly detection.}
% \label{fig:architurcture}
% \end{figure}

\subsection{Graph Representation of Image} 
\begin{figure}[t]
    \centering
    \includegraphics[width=0.85\linewidth]{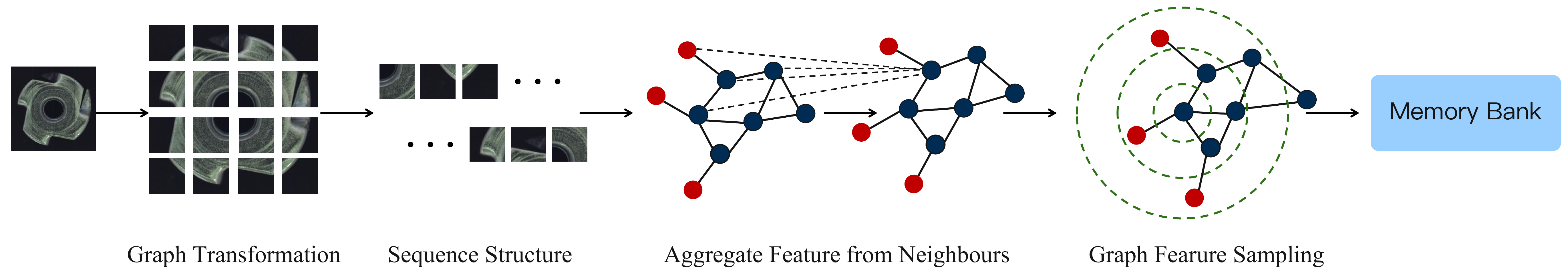}
    \caption{Vision isometric invariant GNN pipeline.}
    \label{fig:gnn_arch}
\end{figure}

Fig.~\ref{fig:gnn_arch} shows the feature extraction process of GraphCore. Specifically, for a normal sample image with a size of $H \times W \times 3$, we evenly separate it as an $N$ patch. In addition, each patch is transformed into a feature vector $f_{i} \in \mathbb{R}^{D}$. So we have the features $F = \left [f_{1}, f_{2},\cdots, f_{N}  \right ]$, where $D$ is the feature dimension and $i=1,2,\cdots,N$. We view these features as unordered nodes $\mathcal{V} = \left\{v_{1}, v_{2}, \cdots, v_{N} \right\}$. For certain each node $v_{i}$, we denote the $K$ nearest neighbours $\mathcal{N}(v_{i})$ and add an edge $e_{ij}$ directed from $v_{j}$ to $v_{i}$ for all $v_{j} \in \mathcal{N}(v_{i})$. Hence, each patch of normal samples can be denoted as a graph $\mathcal{G} = (\mathcal{V},\mathcal{E})$. $\mathcal{E}$ refers all the edges of Graph $\mathcal{G}$.

\begin{figure}[b]
    \centering
    \includegraphics[width=0.7\linewidth]{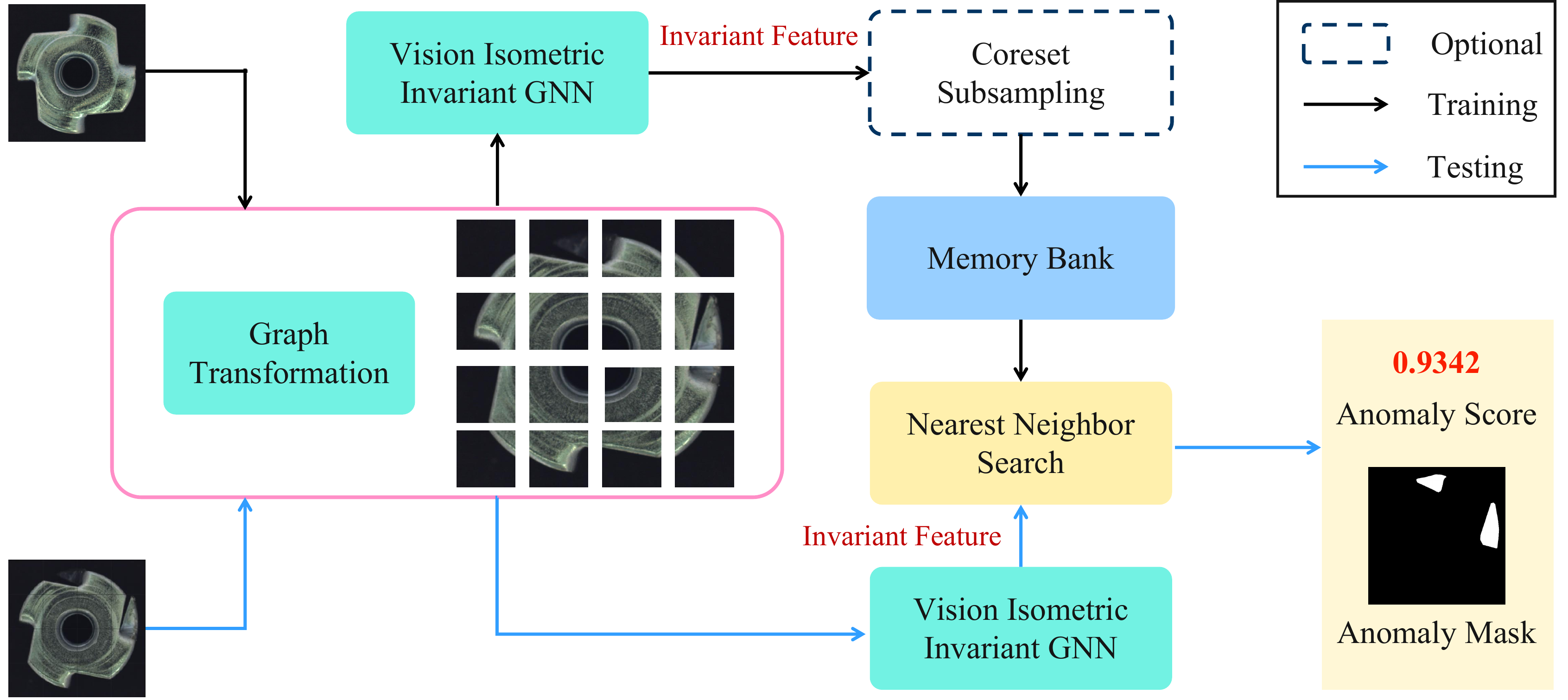}
    \caption{Vision isometric invariant GNN for FSAD.}
    \label{fig:gnn_pipeline}
\end{figure}

\subsection{Graph Feature Processing}
Fig.~\ref{fig:gnn_arch} shows the architecture of the proposed vision isometric invariant GNN. To be specific, we set the feature extraction as GCN (\cite{Kipf2017SemiSupervisedCW}). We aggregate features for each node by exchanging information with its neighbour nodes. In specific, the feature extraction operates as follows:
\begin{equation}
    \mathcal{G}^{'} = F(\mathcal{G}, \mathcal{W}) = Update(Aggregate(\mathcal{G}, W_{aggregate}), W_{update}),
\end{equation}
where $W_{aggregate}$ and $W_{update}$ denote the weights of the aggregation and update operations. Both of them can be optimized in an end-to-end manner. Specifically, the aggregation operation for each node is calculated by aggregating neighbouring nodes' features:
\begin{equation}
    f^{'}_{i} = h(f_{i}, g(f_{i}, \mathcal{N}(f_{i}), W_{aggregate}), W_{update}) ,
\end{equation}
where \xgy{$h$ is the node feature update function and $g$ is the node feature aggregate feature function}. $\mathcal{N}(f^{l}_{i})$ denotes the set of neighbor nodes of $f^{l}_{i}$ \xgy{at the $l$-th layer}. Specifically, we employ max-relative graph convolution (\cite{Li2019CanGG}) as our operator. So $g$ and $h$ are defined as:
\begin{equation} \label{eq:gdot}
    g(\cdot) = f^{''}_{i} = max(\left\{f_{i} - f_{j} | j \in \mathcal{N}(x_{i})\right\}),
\end{equation}
\begin{equation}\label{eq:hdot}
    h(\cdot) = f^{'}_{i} = f^{''}_{i}W_{update}.
\end{equation}
\xgy{In Equations \ref{eq:gdot} and \ref{eq:hdot}, $g(\cdot)$ is a max-pooling vertex feature aggregator that aggregates the difference in features between node $v_i$ and all of its neighbours. $h(\cdot)$ is an MLP layer with batch normalization and ReLU activation.}

\subsection{GraphCore Architecture}

Fig.~\ref{fig:gnn_pipeline} shows the whole architecture of GraphCore.
In the training phase, the most significant difference between GraphCore and Augmentation+PatchCore is the feature memory bank construction algorithm. The feature construction algorithm is the same as Aug.(R) memory bank in Algorithm~\ref{alg:aug}. Note that we use vision isometric invariant GNN as feature extractor $\mathcal{P}$ without data augmentation. In the testing phase, the computation of anomaly score $s*$ for GraphCore is highly similar to the one in Augmentation+PatchCore. The only difference is the feature extraction method for each normal patch sample. The architecture details of the GraphCore are shown in the reference Table~\ref{tab:arch_graphcore}.

% \begin{algorithm}
% \small
% \caption{GraphCore memory bank}\label{alg:graphcore}
%     \SetKwInOut{KwIn}{Input}
%     \SetKwInOut{KwOut}{Output}
%     \KwIn{Vision Isometric GNN $\phi_{G}$, all normal samples $\mathcal{F}_{N}$, memory size target $l$, random linear projection $\psi$.}
%     \KwOut{Patch-level memory bank $\mathcal{M}$.}
%     $\mathcal{M} \leftarrow \left\{ \right\}$\;
%     \For{$f_{i} \in \mathcal{F_{N}}$}
%     {
%         $\mathcal{M} \leftarrow \mathcal{P}(\phi(f_{i}))$\;
%     }
%     $\mathcal{M}_{C} \leftarrow \left\{ \right\}$ \textcolor{gray}{//Apply coreset sampling for memory bank}\\
%     \For{$i \in \left [0,\cdots, l-1  \right ]$}
%     {
%         $m_{i} \leftarrow \underset{m \in \mathcal{M} - \mathcal{M_{C}}}{\argmax} \underset{n \in \mathcal{M_{C}}} {\min} \left\| \psi(m) - \psi(n) \right\|_{2}$\;
%         $\mathcal{M_{C}} \leftarrow \mathcal{M_{C}} \cup \left\{m_{i} \right\}$\;
%     }
%     $\mathcal{M} \leftarrow \mathcal{M_{C}}$.
% \end{algorithm}

\subsection{A Unified View of Augmentation+PatchCore and GraphCore}

%CN:图6给出了统一视角下两种方法，两种方法最大的不同点是在于特征提取的时候， GraphCore是能够得到isometric invariant feature, 所以在memeory bank留下的feature 是更有可能能够在test 阶段用得上
\begin{figure}
    \centering
    \includegraphics[width=0.7\linewidth]{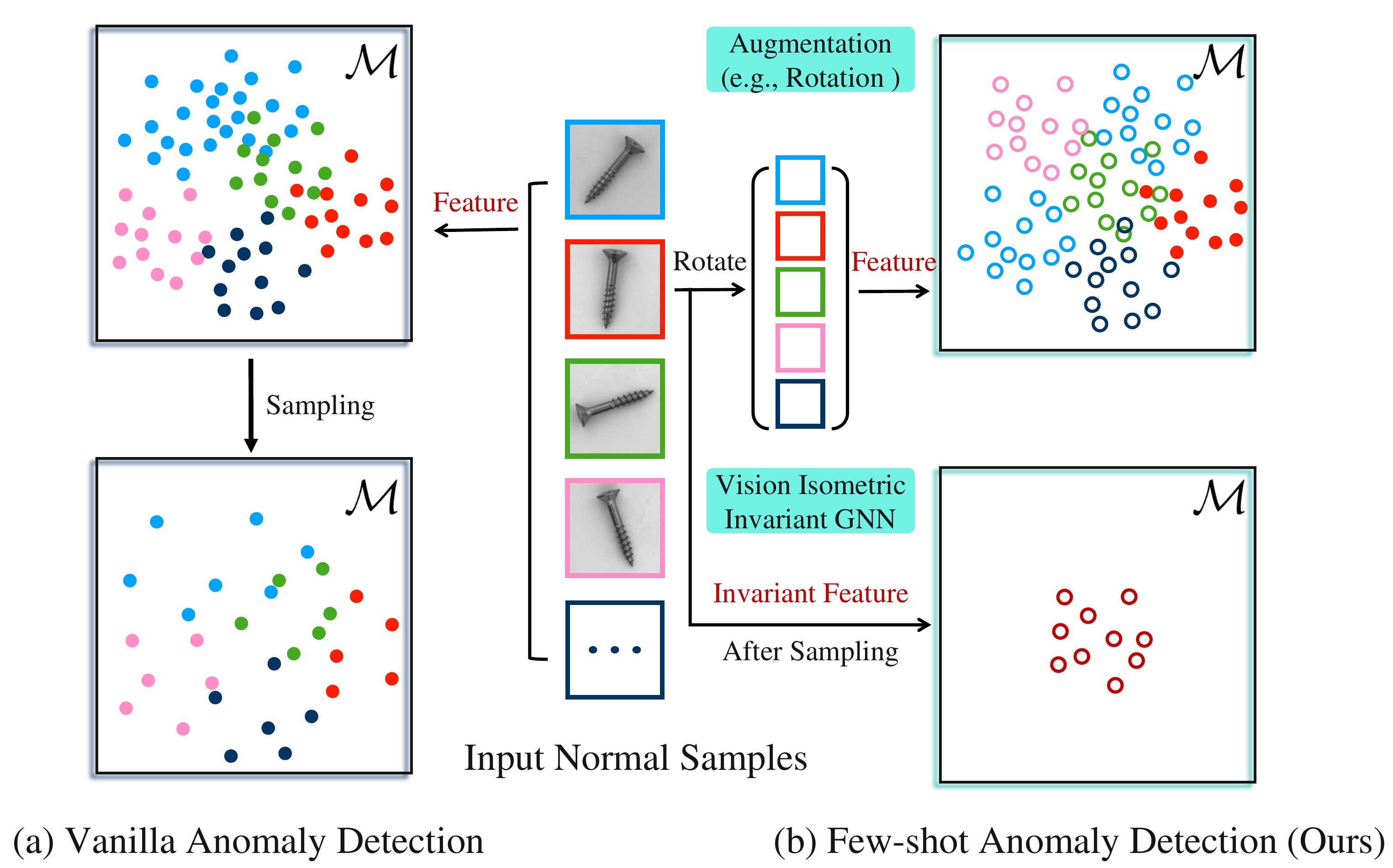}
    \caption{Vision GNN architecture. (a) vanilla AD and (b) our proposed FSAD.}
    \label{fig:unified_view}
\end{figure}

% Fig.~\ref{fig:unified_view} shows a unified view for both Augmentation+PatchCore and GraphCore. Inspired by Augmentation+PatchCore, GraphCore aims to obtain the isometric invariant feature. Hence, GraphCore is able to increase the possibilities to find a feature subset such that the anomaly score of test image can be most accurately and quickly to calculate. From Table~\ref{tab:unified-view-methods}, it is easily observed the difference among PatchCore, Augmentation+PatchCore and GraphCore in terms of architecture details.  
Fig.~\ref{fig:unified_view} depicts a unified view of both Augmentation+PatchCore and GraphCore. Augmentation+PatchCore prompts GraphCore to obtain the isometric invariant feature. Therefore, GraphCore can improve the probability of locating a feature subset, allowing the anomaly score of a test image to be calculated most precisely and rapidly. Table~\ref{tab:unified-view-methods} shows the difference between PatchCore, Augmentation+PatchCore and GraphCore in terms of architectural details. 

\begin{table}[ht]
\caption{Unified view for three methods.}
\centering
\small
% \renewcommand{\arraystretch}{1.2}
% \resizebox{\textwidth}{!}{
    \begin{tabular}{lll}
    \hline
     \textbf{Augmentation} & \textbf{Network} & \textbf{Model} \\
    \hline
     No & ImageNet Pre-trained Model & PatchCore \\
     Rotation & ImageNet Pre-trained Model & Aug.(R) \\
     No & GNN & GraphCore \\
    \hline
    \end{tabular}
% }
\label{tab:unified-view-methods}
\end{table}

\section{Experiment}
%CN:为了证明我们方法的泛化性，我们在三个数据集上选取SOTA的方法在few-shot设定下进行了实验，并将其他相关论文中的实验结果拿来与我们的方法进行比较。
%In order to prove the generalization of our method, we re-implement the SOTA models on 3 datasets under the proposed few-shot settings, and compare the experiment results in other related papers with our method. Due to the page length, we put the hyperparameter setting into the appendix.

\subsection{Experiment setting}
\textbf{Datasets.}
To demonstrate the generalization of our proposed method, we conduct experiments on three datasets, namely MVTec AD (\cite{Bergmann2019MVTecA}), MPDD (\cite{jezek2021deep}) and MVTec LOCO AD (\cite{bergmann2022beyond}).

\textbf{Competing Methods.}
RegAD  (\cite{huang2022registration}) is the SOTA FSAD method. It works under a meta-learning setting: aggregated training on multiple categories and adapting to unseen categories, using few-shot unseen images as a support set. However, our proposed few-shot setting utilizes only a few images as a training set and not several categories. Taking into account the fairness of the experiments, we reimplement the classical and SOTA approaches in the field of unsupervised anomaly detection, such as SPADE (\cite{cohen2020sub}), STPM (\cite{Wang2021StudentTeacherFP}), RD4AD (\cite{Deng2022AnomalyDV}), CFA (\cite{Lee2022CFACF}), and PatchCore (\cite{roth2022towards}), using the official source code for comparison under our few-shot setting. PatchCore-1 is the result of our reimplementation with a 1\% sampling rate, PatchCore-10 and PatchCore-25 are the results at 10\% and 25\% sampling rates, respectively, and RegAD-L is the RegAD experiment with our few-shot setting.

% In addition, we obtain experimental results such as PaDiM (\cite{defard2021padim}), PatchCore-10, PatchCore-25 (\cite{roth2022towards}), RegAD-L (\cite{huang2022registration}) with the same settings from other papers for comparison. 
%  All comparisons are under the same shot number.
%CN:我们选取了当前在FSAD领域SOTA的方法RegAD,在few-shot设定下进行了实验的常规无监督AD方法,并复现了非FSAD领域SOTA的方法在few-shot setting下进行比较。RegAD采用了meta-learnin的设定，而其他方法均在每个种类上都单独进行训练。Aug.(R)为我们在PatchCore的继续上进行了增强的方法，GraphCore是我们针对图像旋转不变性提出的方法。CFA，SPADE,STPM,RD4AD,PatchCore的方法均为我们自己复现，PaDiM,PatchCore-10,PatchCore-25的指标结果来自于 \cite{roth2022towards}，RegAD的指标结果来自于 \cite{huang2022registration}.

%\textbf{Evaluation Metrics.}
%We use the Area Under the Receiver Operating Characteristic (AUROC) as the evaluation metric, which is often used as the performance assessment for AD tasks, to estimate the model performance. Anomaly detection and localization are performed using the image-level AUROC and the pixel-level AUROC respectively.

\subsection{Comparison with the SOTA Methods}
%CN:我们在自己的few-shot设定下和其他AD模型比较了性能，并和meta-learning设定下的RegAD比较了性能。表1和表2分别展示了各种方法在MVTec AD和MPDD数据集上的平均性能。可以看出，我们提出的GraphCore在不同shot下均大幅领先于其他方法，甚至在旋转不变性思想上对PatchCore改进得到的Aug.(R)也对其他方法有较高幅度的领先。显而易见，我们的方法在shot数量较少的时候就能够体现出明显的优势，而随着shot数量增加，优势逐渐缩小，这也说明了我们的方法是针对few-shot场景有特别效果的。

\begin{figure}[t]
\centering
\subfigure[MPDD]{
    \label{fig:num_shot_mpdd}
    \includegraphics[width=0.3\textwidth]{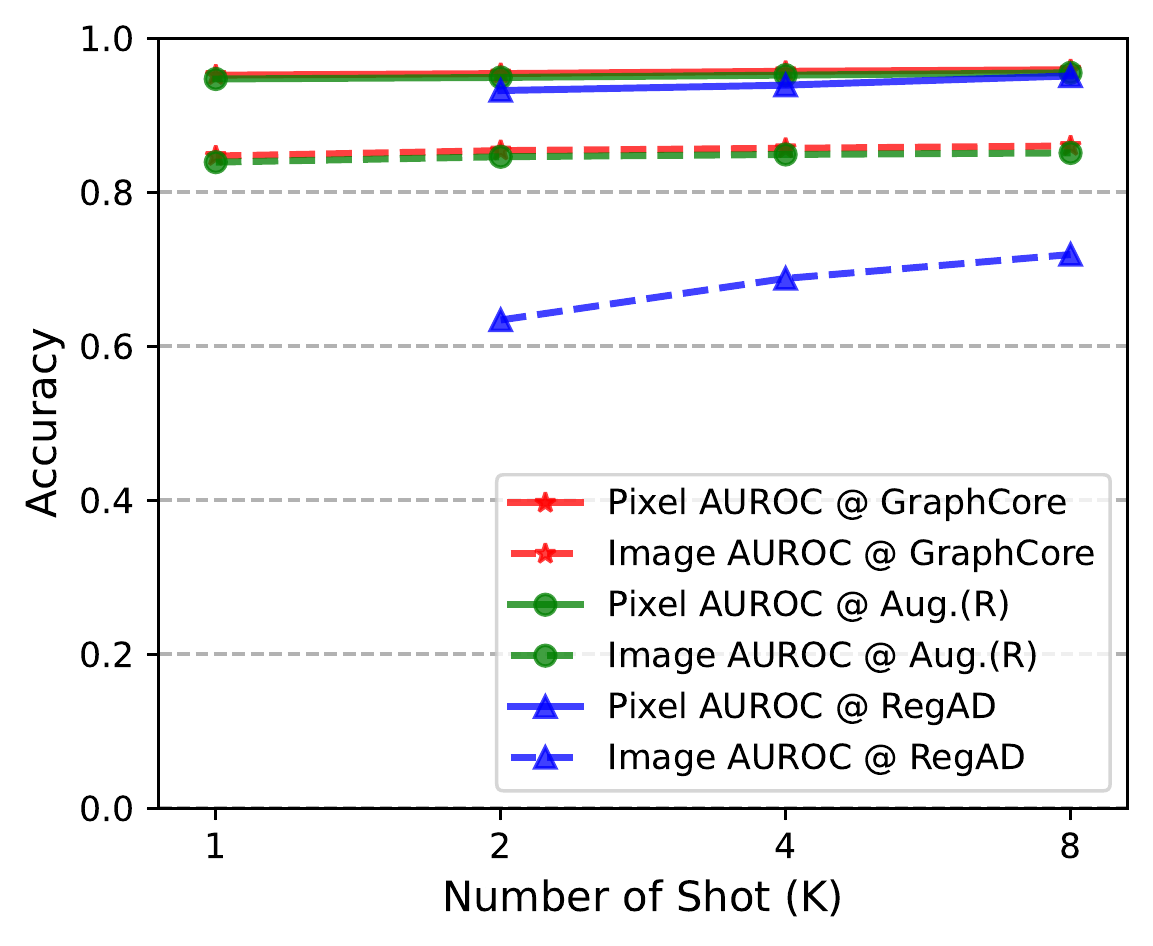}}
        \hspace{5mm}
\subfigure[MVTec AD]{
    \label{fig:num_shot_mvtec2d}
    \includegraphics[width=0.3\textwidth]{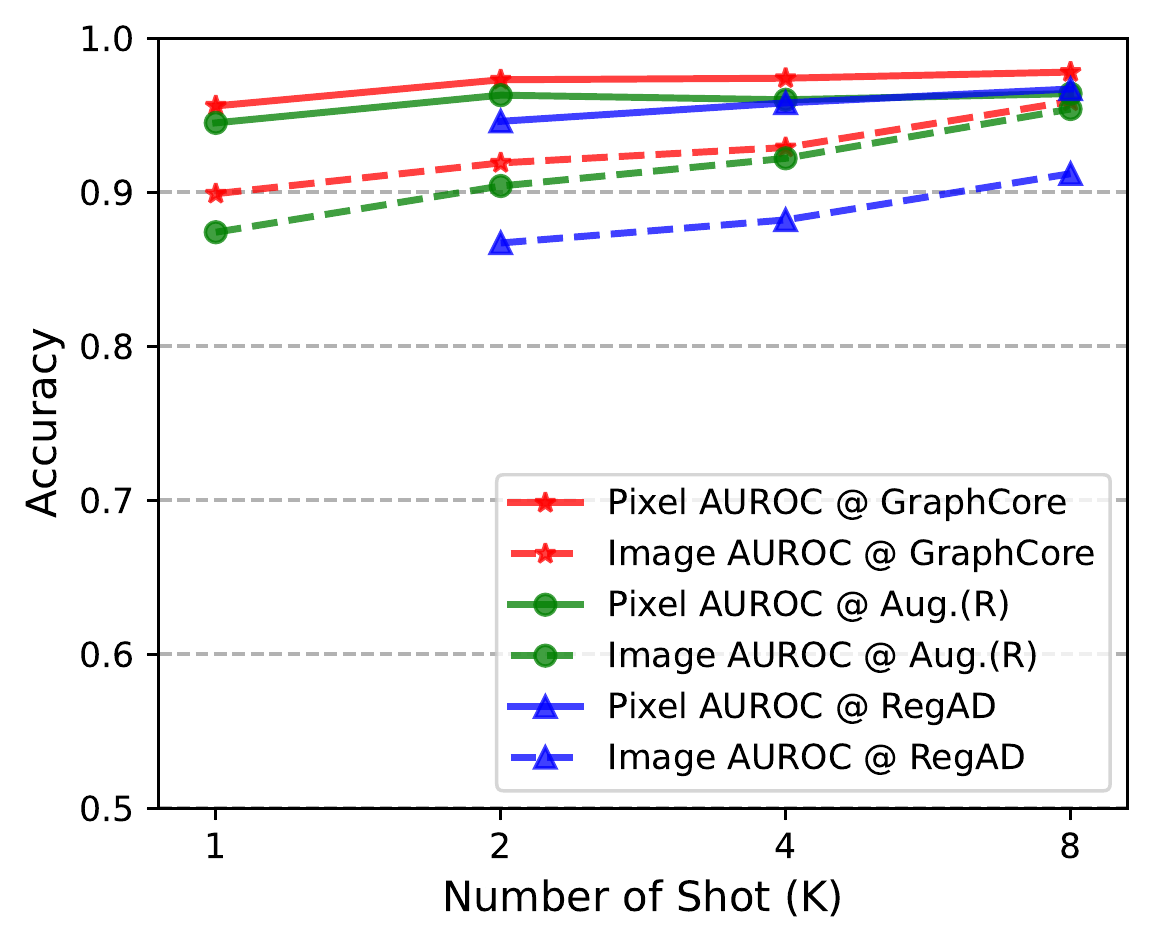}}
\caption{GraphCore VS Augmentation+PatchCore VS RegAD on various numbers of shot (K).}
\label{fig:graph_aug_regad}
\end{figure}

The comparative findings between MVTec and MPDD are shown in Table \ref{tab:total_metric}. Especially the performance of RegAD under the meta-learning setting is also listed in the table. In comparison to SOTA models, GraphCore improves average AUC by 5.8\%, 4.1\%, 3.4\%, and 1.6\% on MVTec and by 25.5\%, 22.0\%, 16.9\%, and 14.1\% on MPDD for 1, 2, 4, and 8-shot cases, respectively. From Fig.~\ref{fig:graph_aug_regad}, it can be easily observed that GraphCore significantly outperforms the SOTA approach at the image and pixel level from 1-shot to 8-shot. 
As can be seen, the performance of GraphCore and Augmentation+PatchCore surpasses the other methods when using only a few samples for training.
%Even our Aug.(R) improvements based on PatchCore are far superior to other methods. Obviously, the result can show obvious advantages of our method when the number of shot is small, and the advantages gradually shrink with the number of shots increases, which also shows that our method has special effects for few-shot scenarios.

\begin{table}[ht]
% \setlength{\abovecaptionskip}{2pt}
%\caption{Results of anomaly detection. (MVTec, K=1 Shot, AUROC)}
\caption{FSAD average results for all categories on MVTec AD and MPDD. The sampling ratio is 0.01, x$|$y represents image AUROC and pixel AUROC. The results for PaDiM, PatchCore-10 and PatchCore-25 are reported from \cite{roth2022towards}. The results for RegAD-L and RegAD are reported from \cite{huang2022registration}. The best-performing method is in bold.}
% \normalsize
\renewcommand{\arraystretch}{1.2}
% \large
\resizebox{\textwidth}{!}{
    % \begin{tabular}{l|l|ll lll lllllllll}
    \begin{tabular}{c|c|cc ccc ccccccc}
    \hline
     \textbf{Dateset} &\textbf{K}& \textbf{Aug.(R)} & \textbf{GraphCore}  & \textbf{CFA} & \textbf{SPADE} & \textbf{PaDiM} & \textbf{STPM} & \textbf{RD4AD} & \textbf{PatchCore-1}& \textbf{PatchCore-10}& \textbf{PatchCore-25} & \textbf{RegAD-L} & \textbf{RegAD} \\
        %  & \textbf{+PatchCore}
        %  \cite{rotation_augmentation} & &~\cite{Lee2022CFACF} &~\cite{cohen2020sub} &~\cite{defard2021padim} &~\cite{sheynin2021hierarchical} &~\cite{Wang2021StudentTeacherFP} &~\cite{Deng2022AnomalyDV} & ~\cite{roth2022towards}& & &~\cite{huang2022registration} \\
     \hline
        \multirow{4}{1.5cm}{MVTec AD} 
        &  1 &  87.4$\mid$94.5 &  \textbf{89.9}$\mid$\textbf{95.6} &  78.8$\mid$90.7 &  69.8$\mid$79.1 &  76.1$\mid$88.2 &  69.7$\mid$58.2 &  74.4$\mid$69.0 &  78.5$\mid$90.1 &  83.4$\mid$92.0 &  84.1$\mid$92.4 &  - &  82.4$\mid$- \\
        &  2 &  90.4$\mid$96.3 &  \textbf{91.9}$\mid$\textbf{96.9} &  81.1$\mid$91.0 &  70.7$\mid$79.9 &  78.9$\mid$90.5 &  74.2$\mid$59.8 &  75.5$\mid$71.8 &  87.8$\mid$94.8 &  86.4$\mid$93.1 &  87.2$\mid$93.3 &  81.5$\mid$- &  85.7$\mid$94.6 \\
        &  4 &  92.2$\mid$96.0 &  \textbf{92.9}$\mid$\textbf{97.4} &  85.0$\mid$91.3 &          - &  71.6$\mid$80.2 &  74.8$\mid$60.8 &  76.9$\mid$72.2 &  89.5$\mid$95.0 &          - &          - &  84.9$\mid$- &  88.2$\mid$95.8 \\
        &  8 &  95.4$\mid$96.4 &  \textbf{95.9}$\mid$\textbf{97.8} &  90.9$\mid$91.6 &          - &  75.3$\mid$80.5 &  77.6$\mid$61.6 &  78.5$\mid$73.0 &  94.3$\mid$95.6 &          - &          - &  87.4$\mid$- &  91.2$\mid$96.7 \\
        \hline
        \multirow{4}{1.5cm}{MPDD} 
        &  1 &  83.9$\mid$94.7 &  \textbf{84.7}$\mid$\textbf{95.2} &  58.8$\mid$77.7 &          - &  57.5$\mid$73.9 &  59.2$\mid$75.1 &  58.5$\mid$73.2 &  59.2$\mid$78.5 &          - &          - &  - &  57.8$\mid$- \\
        &  2 &  84.6$\mid$94.9 &  \textbf{85.4}$\mid$\textbf{95.4} &  58.6$\mid$78.2 &          - &  58.0$\mid$75.4 &  62.4$\mid$75.8 &  61.8$\mid$74.5 &  59.6$\mid$79.2 &          - &          - &  50.8$\mid$- &  63.4$\mid$93.2 \\
        &  4 &  84.9$\mid$95.2 &  \textbf{85.7}$\mid$\textbf{95.7} &  59.3$\mid$78.7 &          - &  58.3$\mid$75.9 &  62.6$\mid$76.2 &  62.1$\mid$75.5 &  59.8$\mid$79.8 &          - &          - &  54.2$\mid$- &  68.8$\mid$93.9 \\
        &  8 &  85.1$\mid$95.5 &  \textbf{86.0}$\mid$\textbf{95.9} &  60.9$\mid$79.0 &          - &  58.5$\mid$76.2 &  63.1$\mid$76.6 &  62.4$\mid$75.7 &  60.0$\mid$80.3 &          - &          - &  61.1$\mid$- &  71.9$\mid$95.1 \\
    \hline
    \end{tabular}
}
\label{tab:total_metric}
\end{table}

Considering that RegAD only shows detailed results of various categories above 2-shot, we only show the detailed results of 2-shot in the main text, and the results of 1-shot, 4-shot, and 8-shot are in the appendix. As shown in Table \ref{tab:mvtec-2shot}, GraphCore outperforms all other baseline methods in 12 out of the 15 categories at the image level and outperforms all other baselines in 11 out of the 15 categories at the pixel level on MVTec AD. Moreover, results in Table \ref{tab:mpdd-2shot} show that GraphCore outperforms all other baselines in 5 out of the 6 categories at the image level and outperforms all other baselines in all categories at the pixel level on MPDD.

\begin{table}[ht]
% \setlength{\abovecaptionskip}{2pt}
%\caption{Results of anomaly detection. (MVTec, K=1 Shot, AUROC)}
\caption{FSAD results on MVTec AD. The number of shots K is 2, and the sampling ratio is 0.01, x$|$y represents image AUROC and pixel AUROC. The results for PaDiM, PatchCore-10 and PatchCore-25 are reported from \cite{roth2022towards}. The results for RegAD are reported from \cite{huang2022registration}. The best-performing method is in bold.}
\renewcommand{\arraystretch}{1.2}
% \large
\resizebox{\textwidth}{!}{
    % \begin{tabular}{l|ll lll llllll}
    \begin{tabular}{c|c|cc ccc ccccccc}
    \hline
     \textbf{Category} & \textbf{Aug.(R)} & \textbf{GraphCore}  & \textbf{CFA} & \textbf{SPADE} & \textbf{PaDiM} & \textbf{STPM} & \textbf{RD4AD}  & \textbf{PatchCore-1}& \textbf{PatchCore-10}& \textbf{PatchCore-25} & \textbf{RegAD} \\
     \hline
        Bottle     & 99.7$\mid$98.6 & \textbf{99.8$\mid$99.8} & 93.7$\mid$93.5  & 95.7$\mid$86.8 & - & 93.8$\mid$84.6 & 91.2$\mid$81.7 & 99.7$\mid$98.5 & - & - & 99.4$\mid$98.0 \\
        Cable      & 94.7$\mid$96.2 & \textbf{95.2}$\mid$96.3 & 89.3$\mid$88.9  & 60.4$\mid$78.6 & - & 60.2$\mid$51.6 & 65.3$\mid$65.4 & 94.9$\mid$\textbf{97.8} & - & - & 65.1$\mid$91.7 \\
        Capsule    & 66.5$\mid$97.7 & \textbf{73.2}$\mid$\textbf{97.8} & 53.4$\mid$85.9  & 48.7$\mid$79.8 & - & 45.2$\mid$59.2 & 50.5$\mid$78.2 & 67.2$\mid$97.7 & - & - & 67.5$\mid$97.3 \\
        Carpet     & \textbf{99.4}$\mid$99.1 & \textbf{99.4}$\mid$\textbf{99.6} & 97.6$\mid$97.9  & 92.1$\mid$95.6 & - & 90.8$\mid$60.5 & 92.8$\mid$74.2 & 99.1$\mid$99.0 & - & - & 96.5$\mid$98.9 \\
        Grid       & 75.7$\mid$79.8 & 81.5$\mid$80.6 & 80.4$\mid$\textbf{81.4}  & 75.8$\mid$75.9 & - & 72.6$\mid$61.2 & 75.2$\mid$76.3 & 61.7$\mid$67.5 & - & - & \textbf{84.0}$\mid$77.4 \\
        Hazelnut   & \textbf{99.7}$\mid$97.9 & 99.5$\mid$\textbf{98.2} & 99.4$\mid$\textbf{98.2}  & 95.2$\mid$88.9 & - & 90.3$\mid$74.5 & 93.4$\mid$64.8 & 93.5$\mid$96.4 & - & - & 96.0$\mid$98.1 \\
        Leather    & \textbf{100}$\mid$99.3  & \textbf{100}$\mid$\textbf{99.4}  & \textbf{100}$\mid$99.3   & 97.9$\mid$89.2 & - & 95.8$\mid$75.2 & 96.7$\mid$86.5 & \textbf{100}$\mid$99.3  & - & - & 99.4$\mid$98.0 \\
        Meta Nut   & 95.0$\mid$96.8 & \textbf{96.3}$\mid$\textbf{98.1} & 68.6$\mid$89.7  & 61.3$\mid$59.5 & - & 59.4$\mid$51.1 & 63.4$\mid$68.9 & 92.0$\mid$97.1 & - & - & 91.4$\mid$96.9 \\
        Pill       & 87.8$\mid$93.9 & \textbf{88.6}$\mid$94.1 & 67.4$\mid$91.5  & 60.2$\mid$57.2 & - & 58.7$\mid$49.9 & 62.8$\mid$70.2 & 87.4$\mid$\textbf{96.8} & - & - & 81.3$\mid$93.6 \\
        Screw      & 63.6$\mid$96.0 & \textbf{65.7}$\mid$96.5 & 58.2$\mid$\textbf{96.7}  & 51.3$\mid$70.2 & - & 51.9$\mid$51.8 & 54.3$\mid$60.8 & 48.3$\mid$90.8 & - & - & 52.5$\mid$94.4 \\
        Tile       & \textbf{100}$\mid$99.3  & \textbf{100}$\mid$\textbf{96.8}  & 99.8$\mid$81.8  & 90.2$\mid$82.3 & - & 91.4$\mid$58.2 & 88.9$\mid$59.2 & \textbf{100}$\mid$96.0  & - & - & 94.3$\mid$94.3 \\
        Toothbrush & 83.6$\mid$98.2 & \textbf{87.3}$\mid$\textbf{98.6} & 86.9$\mid$93.9  & 80.2$\mid$76.8 & - & 76.5$\mid$66.3 & 77.1$\mid$78.3 & 83.9$\mid$98.2 & - & - & 86.6$\mid$98.2 \\
        Transistor & 96.3$\mid$94.1 & \textbf{97.1}$\mid$\textbf{99.2} & 72.5$\mid$80.3  & 51.6$\mid$73.6 & - & 82.4$\mid$47.5 & 78.1$\mid$67.8 & 96.9$\mid$95.0 & - & - & 86.0$\mid$93.4 \\
        Wood       & 97.1$\mid$98.4 & 97.5$\mid$\textbf{99.5} & 98.2$\mid$92.4  & 50.3$\mid$89.7 & - & 95.8$\mid$48.4 & 93.7$\mid$93.8 & 97.2$\mid$93.0 & - & - & \textbf{99.2}$\mid$93.5 \\
        Zipper     & 96.9$\mid$99.0 & \textbf{97.5}$\mid$\textbf{99.3} & 50.5$\mid$94.1  & 49.4$\mid$93.7 & - & 47.6$\mid$56.3 & 49.5$\mid$51.2 & 95.3$\mid$98.2 & - & - & 86.3$\mid$95.1 \\
        \hline
        Average    & 90.4$\mid$96.3 & \textbf{91.9}$\mid$\textbf{96.9} & 81.1$\mid$91.0 & 70.7$\mid$79.9 & 78.9$\mid$90.5 & 74.2$\mid$59.8 & 75.5$\mid$71.8 & 87.8$\mid$94.8 & 86.4$\mid$93.1 & 87.2$\mid$93.3 & 85.7$\mid$94.6 \\
    \hline
    \end{tabular}
}
\label{tab:mvtec-2shot}
\end{table}

\begin{table}[ht]
% \setlength{\abovecaptionskip}{2pt}
%\caption{Results of anomaly detection. (MVTec, K=1 Shot, AUROC)}
\caption{FSAD results on MPDD. The number of shots K is 2, and the sampling ratio is 0.01, x$|$y represents image AUROC and pixel AUROC. The results for PaDiM, PatchCore-10 and PatchCore-25 are reported from \cite{roth2022towards}. The results for RegAD are reported from \cite{huang2022registration}. The best-performing method is in bold.}
% \normalsize
\renewcommand{\arraystretch}{1.2}
% \large
\resizebox{\textwidth}{!}{
    % \begin{tabular}{l|ll lll lll}
    \begin{tabular}{c|c|cc ccc ccccccc}
    \hline
     \textbf{Category} & \textbf{Aug.(R)} & \textbf{GraphCore}  & \textbf{CFA} & \textbf{SPADE} & \textbf{STPM} & \textbf{RD4AD}  & \textbf{PatchCore} & \textbf{RegAD} \\
        %  & \textbf{+PatchCore} & &~\cite{Lee2022CFACF} &~\cite{cohen2020sub} &~\cite{Wang2021StudentTeacherFP} &~\cite{Deng2022AnomalyDV} & ~\cite{roth2022towards}&~\cite{huang2022registration}  \\
     \hline
        Bracket Black & 66.8$\mid$92.1 & 67.0$\mid$\textbf{92.5} & 54.3$\mid$75.8  & 62.4$\mid$72.8  & \textbf{94.5}$\mid$75.1 & 91.7$\mid$75.4 & 58.6$\mid$78.9 &  63.3$\mid$- \\
        Bracket Brown & 76.1$\mid$91.9 & \textbf{77.2}$\mid$\textbf{92.6} & 66.8$\mid$77.5  & 59.5$\mid$71.9  & 62.3$\mid$73.2 & 58.8$\mid$73.4 & 70.7$\mid$76.9 &  59.4$\mid$- \\
        Bracket White & 87.2$\mid$97.1 & \textbf{89.4}$\mid$\textbf{97.5} & 68.7$\mid$70.8  & 67.2$\mid$72.4  & 53.8$\mid$64.2 & 55.6$\mid$62.4 & 70.4$\mid$68.1 &  55.6$\mid$- \\
        Connector     & 98.6$\mid$97.2 & \textbf{98.9}$\mid$\textbf{97.7} & 58.5$\mid$88.2  & 59.2$\mid$82.8  & 51.6$\mid$83.4 & 53.7$\mid$82.3 & 59.2$\mid$85.2 &  73.0$\mid$- \\
        Metal Plate   & \textbf{99.9}$\mid$98.4 & \textbf{99.9}$\mid$\textbf{99.1} & 62.7$\mid$84.3  & 64.2$\mid$75.9  & 62.4$\mid$83.2 & 65.2$\mid$76.5 & 64.1$\mid$86.3 &  61.7$\mid$- \\
        Tubes         & 79.2$\mid$92.6 & \textbf{79.8}$\mid$\textbf{93.1} & 40.7$\mid$72.8  & 35.6$\mid$76.8  & 49.6$\mid$75.6 & 45.9$\mid$77.1 & 34.3$\mid$79.5 &  67.1$\mid$- \\
        \hline
        Average       & 84.6$\mid$94.9 & \textbf{85.4}$\mid$\textbf{95.4} & 58.6$\mid$78.2 & 58.0$\mid$75.4  & 62.4$\mid$75.8 & 61.8$\mid$74.5 & 59.6$\mid$79.2  & 63.4$\mid$93.2 \\
    \hline
    \end{tabular}
}
\label{tab:mpdd-2shot}
\end{table}

\subsection{Ablation Studies}
%CN:我们在MVTec AD,MPDD,MVTec LOCO AD三个数据集上，分别针对邻居个数，采样率进行了消融实验。并对PatchCore,Aug.(R),GraphCore在不同shot数下的表现进行了分析。
%We conducted ablation studies on various number of neighbors and the sampling rate. 

\begin{figure}[htb]
\centering
\subfigure[MPDD]{
    \label{fig:samplerate_mpdd}
    \includegraphics[width=0.3\textwidth]{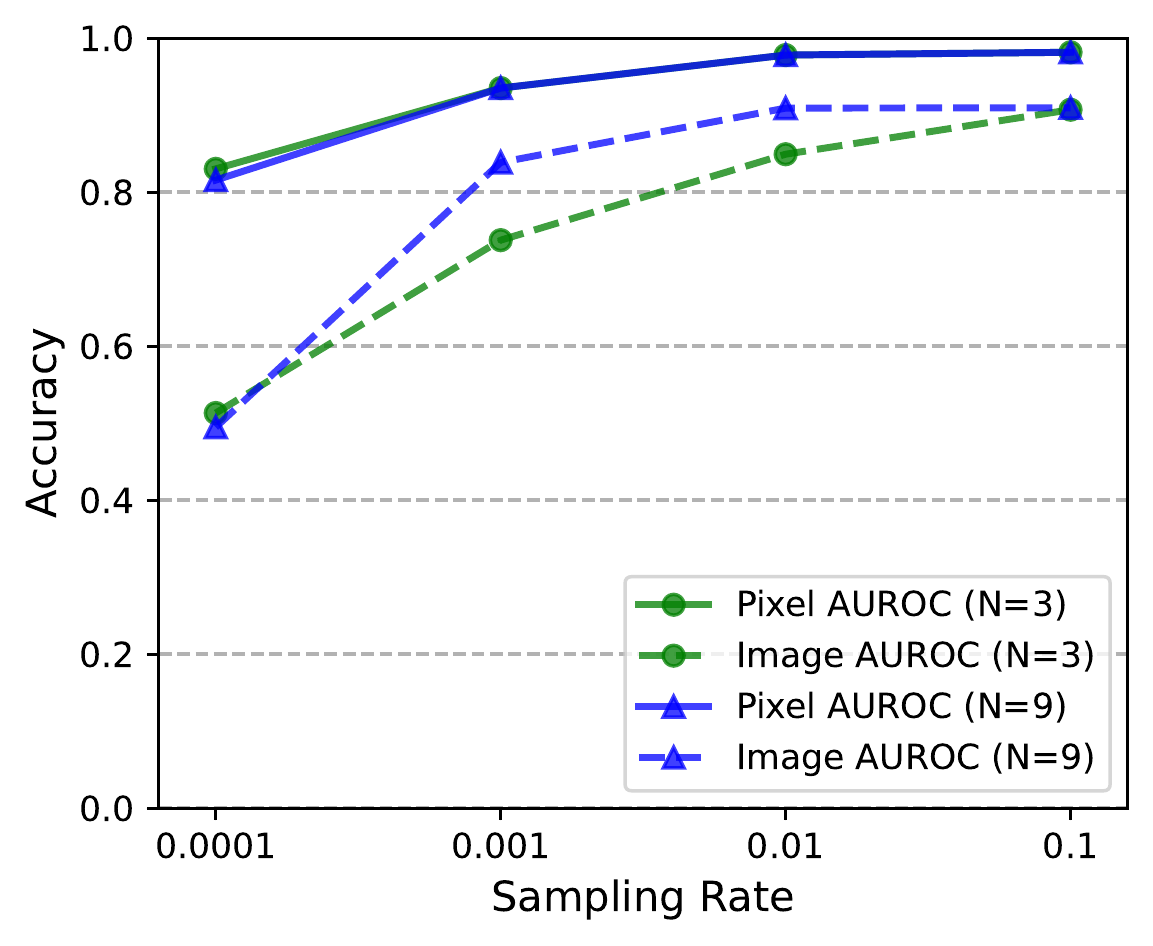}}
        \hspace{-2mm}
\subfigure[MVTec AD]{
    \label{fig:samplerate_mvtec2d}
    \includegraphics[width=0.3\textwidth]{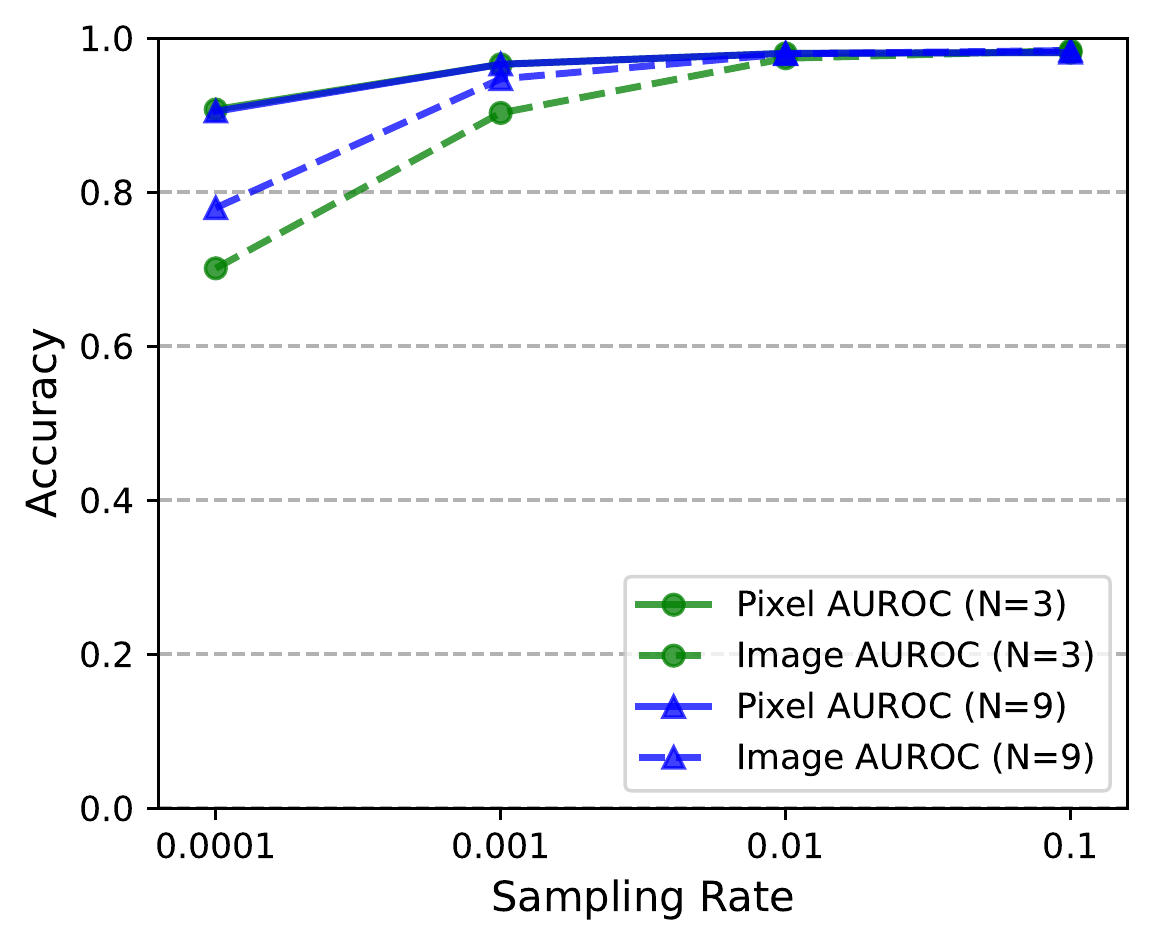}}
        \hspace{-2mm}
\subfigure[MVTec LOCO AD]{
    \label{fig:samplerate_mvteclogical}
    \includegraphics[width=0.3\textwidth]{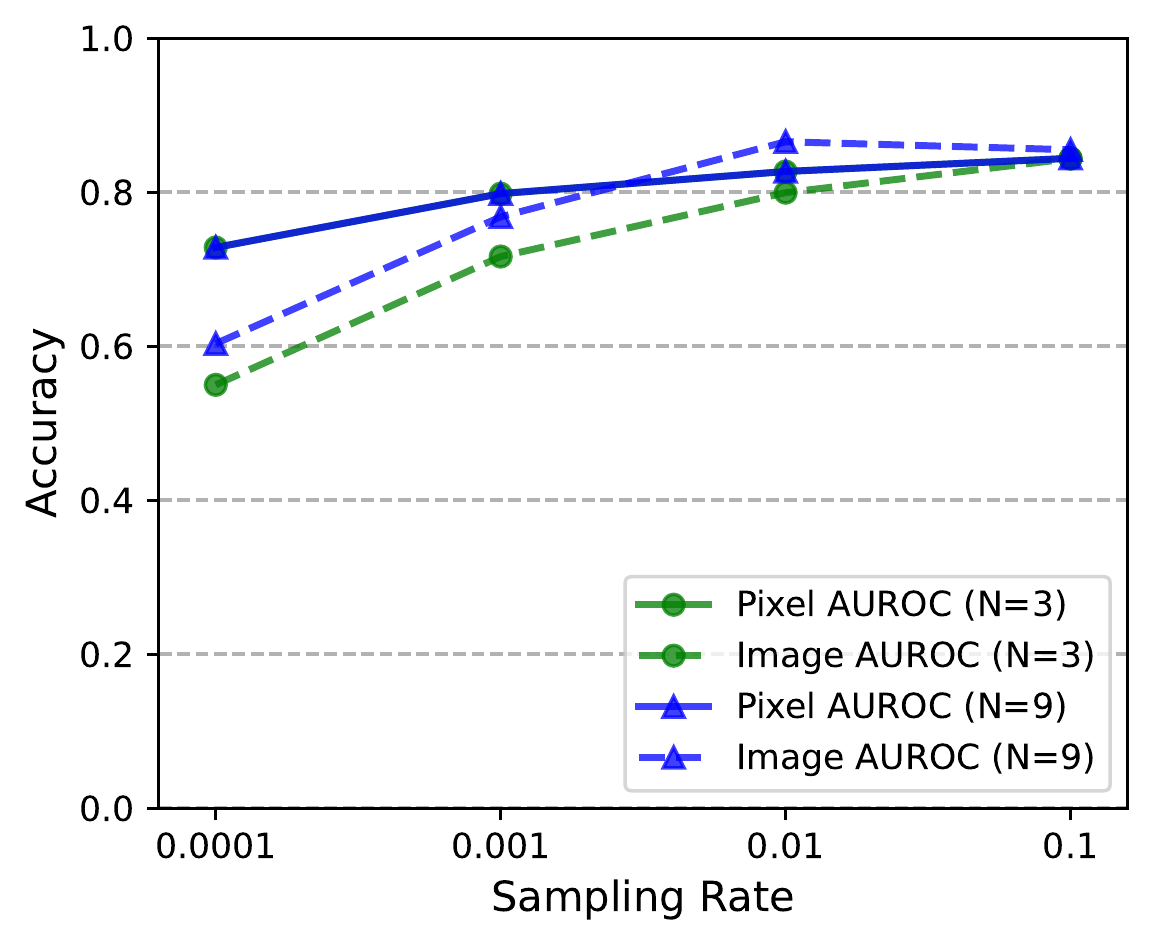}}
\caption{Ablation results on sampling rates and the number of $N$ nearest neighbors.}
\label{fig:ablation_samplingrate}
\end{figure}

\begin{figure}[htb]
\centering
\subfigure[MPDD]{
    \label{fig:num_shot_mpdd}
    \includegraphics[width=0.3\textwidth]{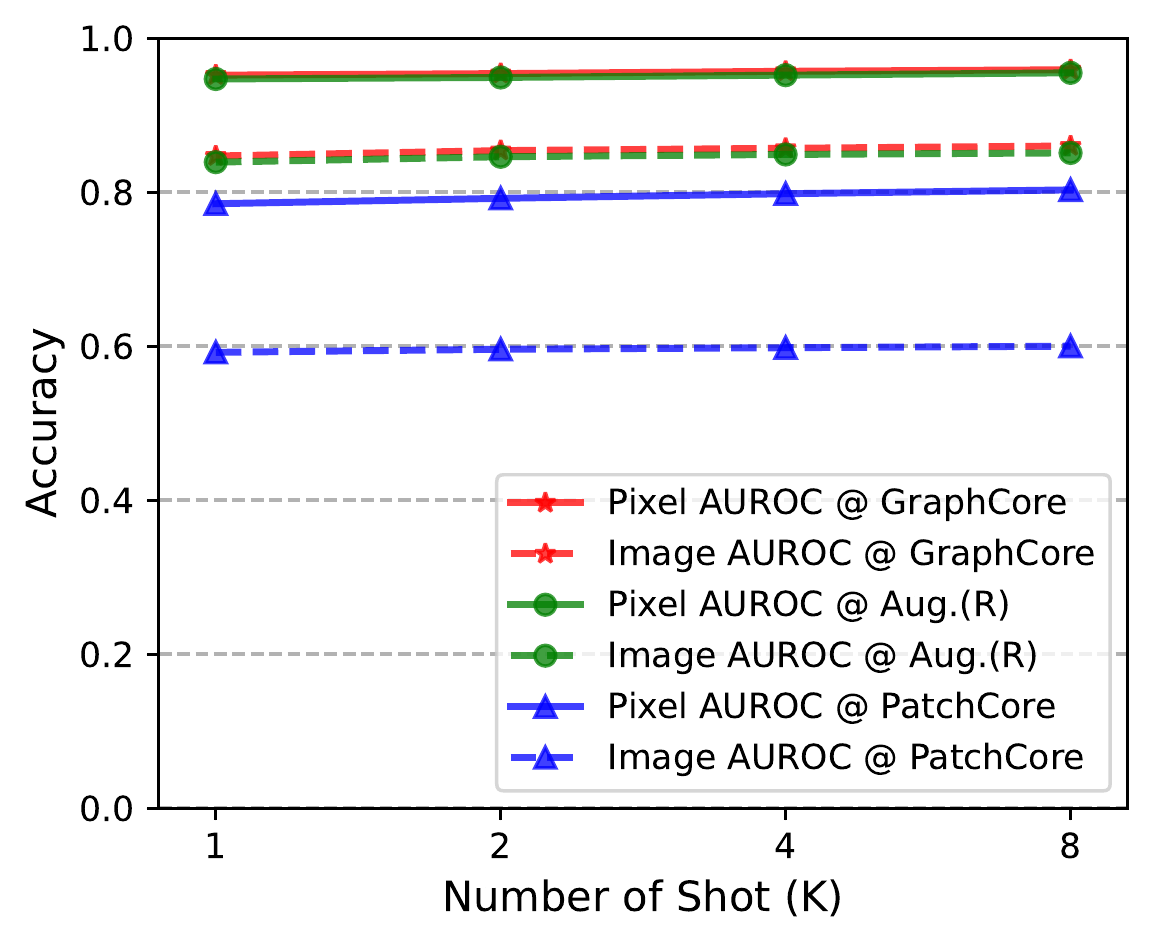}}
        \hspace{5mm}
\subfigure[MVTec AD]{
    \label{fig:num_shot_mvtec2d}
    \includegraphics[width=0.3\textwidth]{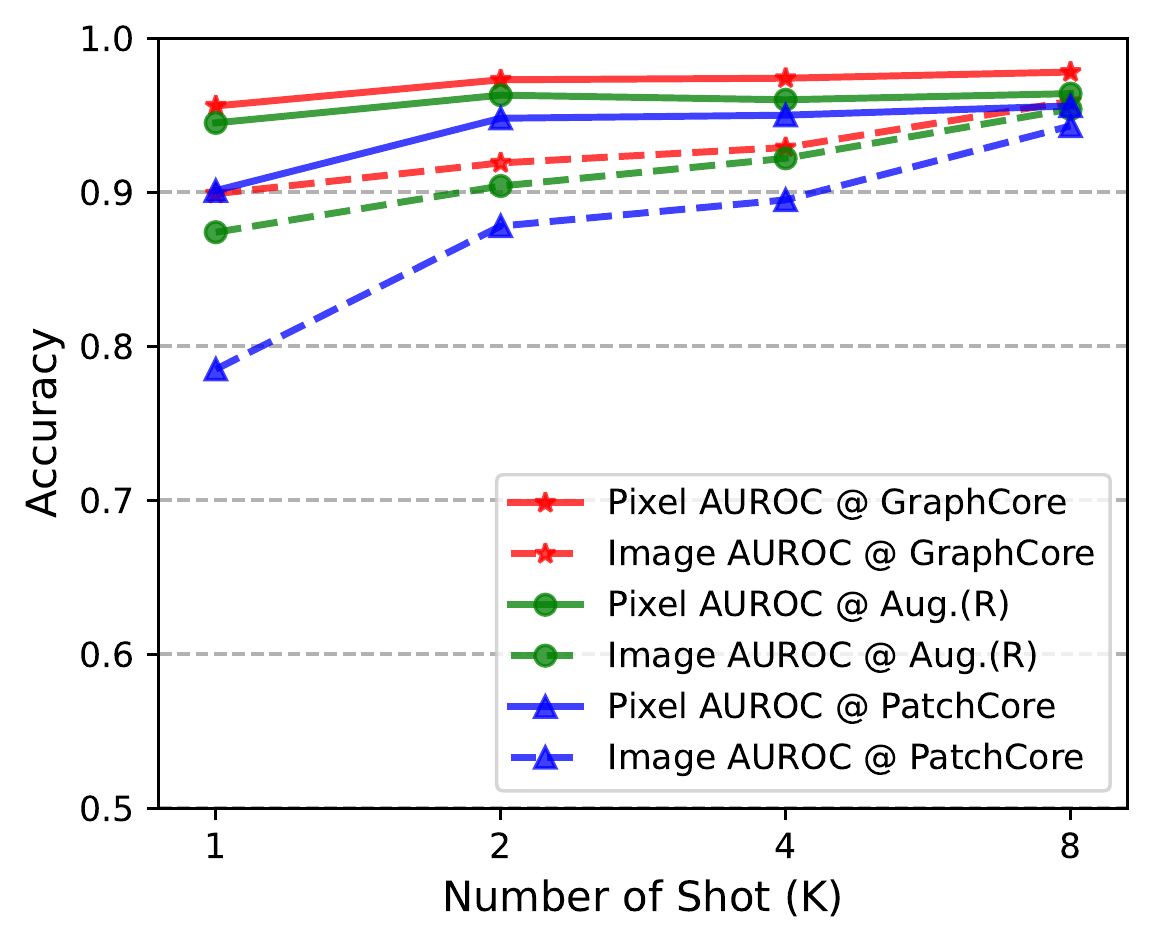}}
\caption{GraphCore vs Augmentation+PatchCore vs PatchCore on various number of shot (K).}
\label{fig:graph_aug_patchcore}
\end{figure}

\textbf{Sampling Rate.}\label{sec:sampling}
%CN:如图所示，我们的方法在采样率从0.0001提升到0.001的时候有较高的提升，随后采样率的提升对性能的增益趋于平缓。
When demonstrated in Fig.~\ref{fig:ablation_samplingrate}, our technique significantly improves as the sample rate increases from 0.0001 to 0.001, after which the increase in sampling rate has a flattening effect on the performance gain. In other words, as the sampling rate steadily increases, the performance of GraphCore is insensitive to the sampling rate.

\textbf{Nearest Neighbour.}
%CN:从图可以看出，将邻居数量从3增加到9在采样较低时对于Pixel level的提升比较明显，但对于image level没有提升。而随着采样率逐渐增大，邻居数量在pixel level上的增益也趋近于无。
In Fig.~\ref{fig:ablation_samplingrate}, the green colour represents the performance of GraphCore's 9 nearest neighbour search, and the blue colour represents the performance of GraphCore's 3 nearest neighbour search. As can be seen, increasing the number of neighbours from 3 to 9 greatly increases performance at the pixel level when the sampling rate is low, but does not enhance performance at the image level. As the sampling rate increases, the gain of the number of pixels' neighbours approaches zero.

\textbf{Augmentation Methods.}
%CN:如图11所示,PatchCore在MVTec AD和MPDD上的性能都相对较低，Aug.(R)则相对PatchCore体现出了卓越的性能，说明了我们对特征的旋转增强是著有成效的。而GraphCore利用GNN抽取特征的能力在这一方面优于特征增强。PatchCore在MVTec AD数据集上随着shot数增加性能逐渐提升,而我们方法的性能在MVTec AD和MPDD上的增长则更加平缓————在shot较少时即达到较高性能水平，这一表现也佐证了旋转不变性对于FSAD的作用。
Fig. \ref{fig:graph_aug_patchcore} demonstrates that the performance of PatchCore on MVTec AD and MPDD is relatively weak, but Aug.(R) demonstrates higher performance. It demonstrates heuristically that our enhancement to feature rotation is significantly effective. Moreover, GraphCore outperforms Aug.(R) by a large margin, confirming our assumption that GraphCore can extract the isometric invariant feature from industrial-based anomaly detection images.

%While GraphCore outperforms PatchCore and \textbf{Aug.(R)} since the ability of GraphCore to use GNN to extract features is superior to feature enhancement in this regard. The performance of PatchCore on the MVTec AD dataset increases gradually with the increase of the number of shots, while the performance of our method on MVTec AD and MPDD increases more smoothly—a higher performance level is achieved when there are fewer shots. The performance also supports the effect of rotational invariance on FSAD.

\subsection{Visualization}
\begin{figure}
    \centering
    \includegraphics[width=0.9\linewidth]{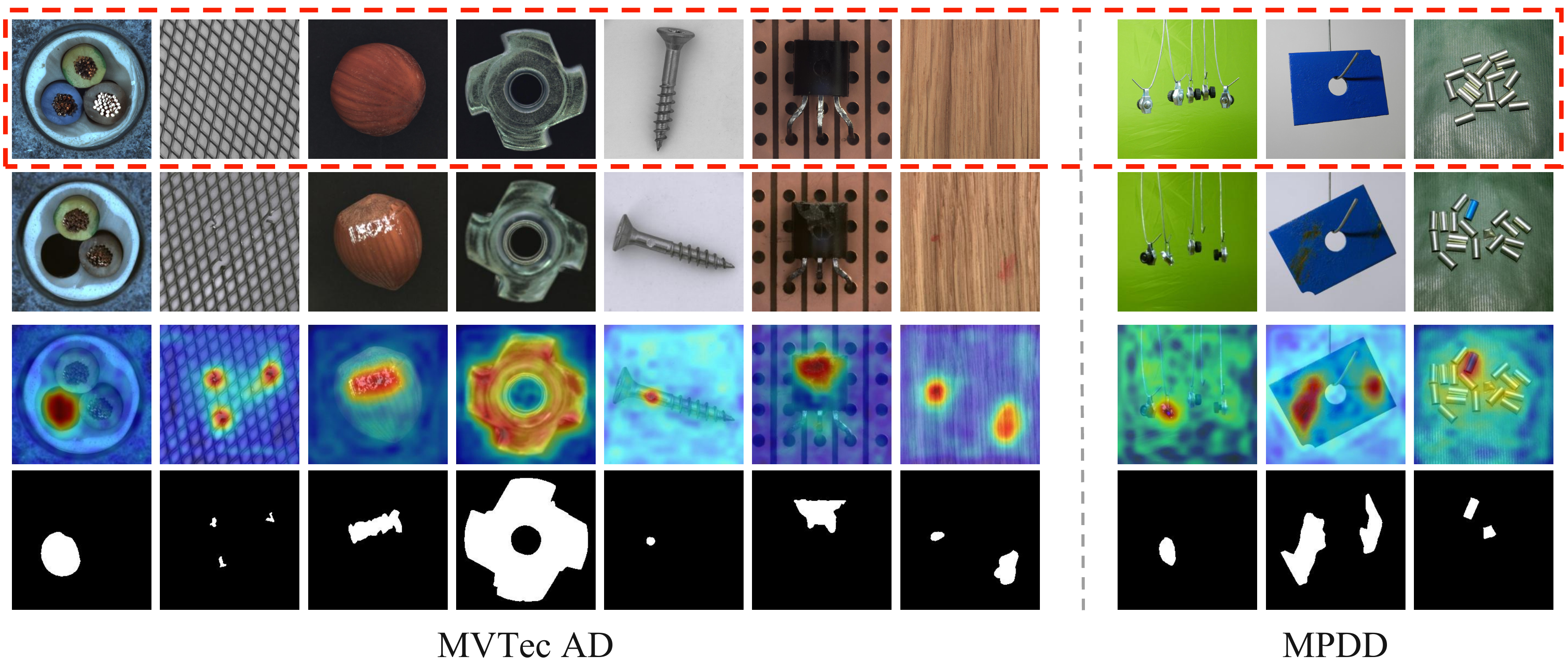}
    \caption{Visualization results of the proposed method on MVTec AD and MPDD. \xgy{The first row denotes the training example in the 1-shot setting. The second row is test samples (abnormal). The third row is the heatmap on test samples. The fourth row is the anomaly mask (ground truth).}}
    \label{fig:visualization}
\end{figure}

%CN:图9和图10分别是我们的方法在0.01的采样率和1shot下在MVTec AD和MPDD数据集上取得的结果。每列代表不同种类物品，四行图像从上到下依次是检测图像，异常得分图，anomaly map on origin image,GT。据图可知我们的方法在各类物体上均可达到较好地异常定位效果，显示它即使是在1-shot的情景下也具有较好的泛化性。
Fig.~\ref{fig:visualization}  shows the visualization results obtained by our method on MVTec AD and MPDD with sampling rates of 0.01 and 1 shot, respectively. Each column represents a different item type, and the four rows, from top to bottom, are the detection image, anomaly score map, anomaly map on detection image, and ground truth. According to the results, our method can produce a satisfactory impact of anomaly localization on various objects, indicating that it has a strong generalized ability even in the 1-shot case.

% \begin{figure}
%     \centering
%     \includegraphics[width=1\linewidth]{iclr2023/fig/visualization.pdf}
%     \caption{Visualization results (MVTec AD) of the proposed method.}
%     \label{fig:visualization}
% \end{figure}

% \begin{figure}
%     \centering
%     \includegraphics[width=0.5\linewidth]{iclr2023/fig/visualization_mpdd.pdf}
%     \caption{Visualization results (MPDD) of the proposed method.}
%     \label{fig:visualization_mpdd}
% \end{figure}

\section{Conclusion}
In this study, we introduce a new approach, GraphCore, for industrial-based few-shot visual anomaly detection. Initially, by investigating the CNN-generated feature space, we present a simple pipeline - Augmentation+PatchCore - for obtaining rotation-invariant features. It turns out that this simple baseline can significantly improve anomaly detection performance. We further propose GraphCore to capture the isometric invariant features of normal industrial samples. It outperforms the SOTA models by a large margin using only a few normal samples ($\leq 8$) for training. The majority of industrial datasets for anomaly detection possess isomorphism, which is a property ideally suited to GraphCore. We will continue to push the limits of industrial-based few-shot anomaly detection in the future.

\section{Acknowledgments} 
This work is supported by the National Natural Science Foundation of China under Grant No. 62122035, 62206122, and 61972188. Y. Jin is supported by an Alexander von Humboldt Professorship for AI endowed by the German Federal Ministry of Education and Research.

\bibliography{iclr2023_conference}
\bibliographystyle{iclr2023_conference}

\section{Appendix}
\subsection{Dataset}
\textbf{MVTec AD} is the most popular dataset for industrial image anomaly detection (\cite{Bergmann2019MVTecA}), which consists of 15 categories of items, including a total of 3629 normal images as a training set, and a collection of 1725 normal images and abnormal images as a test set. All images have a resolution between 700$\times$700 and 1024$\times$1024 pixels.

\textbf{MPDD} is a more challenging AD dataset containing 6 classes of metal parts (\cite{jezek2021deep}). The images are taken in different spatial directions and distances and under the condition of non-uniform background, so it is more challenging. The training set contains 888 normal images, and the test set contains 176 normal images and 282 abnormal images. The resolution of all images is 1024$\times$1024 pixels.

\textbf{MVTEC LOCO AD} adds logical abnormal images outside the structural class abnormal image (\cite{bergmann2022beyond}). The dataset contains 1,772 normal images as a training set, and 304 normal images are used as a validation set. The test set contains 575 normal images, 432 structural abnormal images, and 561 logic abnormal images. Due to the different calculation methods of logic abnormal detection metric, we abandon the logical abnormal image of the test concentration, retaining the remaining 575 normal images and 432 structural abnormal images as a test set for experiments. Each image is 850 to 1600 pixels in height and 800 to 1700 pixels wide.

\subsection{Experiment results}
% Setting: New Few-shot Setting, K (number of shot)=1, dataset: mvtec， sampling ratio: 0.01, metrics: Image AUROC
% *RegAD denote the setting is meta learning setting for few-shot anomaly detection, which is described in Fig. 2, the number of shot in RegAD is 2
\begin{table}[ht]
% \setlength{\abovecaptionskip}{2pt}
%\caption{Results of anomaly detection. (MVTec, K=1 Shot, AUROC)}
\caption{Results of anomaly detection. Setting: New Few-shot Setting, K (number of shot)=1, Dataset: MVTec, Sampling Ratio: 0.01, Metrics: Image AUROC. The number of shot for RegAD is 2. The data for PaDiM and PatchCore-10, PatchCore-25 are from \cite{roth2022towards}.}
% \normalsize
\renewcommand{\arraystretch}{1.2}
% \large
\resizebox{\textwidth}{!}{
    \begin{tabular}{l|ll lll lllllll}
    \hline
     \textbf{Category} & \textbf{Aug.(R)} & \textbf{GraphCore}  & \textbf{CFA} & \textbf{SPADE} & \textbf{PaDiM}  & \textbf{STPM} & \textbf{RD4AD} & \textbf{PatchCore-1} & \textbf{PatchCore-10} & \textbf{PatchCore-25} & \textbf{RegAD} \\
        %  & \textbf{+PatchCore} & &~\cite{Lee2022CFACF} &~\cite{cohen2020sub} &~\cite{defard2021padim} &~\cite{sheynin2021hierarchical} &~\cite{Wang2021StudentTeacherFP} &~\cite{Deng2022AnomalyDV} & ~\cite{roth2022towards}& & &~\cite{huang2022registration}  \\
     \hline
        \large Bottle     & \large 99.7 & \large \textbf{99.8} & \large 96.7  & \large 95.2 & - & \large 93.2 & \large 91.2 & \large 96.5 & - & - & -\\
        \large Cable      & \large 90.1 & \large \textbf{91.1} & \large 65.4  & \large 60.1 & - & \large 59.8 & \large 58.3 & \large 65.5 & - & - & - \\
        \large Capsule    & \large 64.7 & \large \textbf{72.1} & \large 50.2  & \large 45.6 & - & \large 43.2 & \large 44.7 & \large 49.8 & - & - & -\\
        \large Carpet     & \large \textbf{99.3} & \large \textbf{99.3} & \large 97.1  & \large 93.2 & - & \large 90.5 & \large 92.5 & \large 97.2 & - & - & - \\
        \large Grid       & \large 70.8 & \large \textbf{80.9} & \large 79.2  & \large 75.1 & - & \large 71.2 & \large 74.3 & \large 78.9 & - & - & -\\
        \large Hazelnut   & \large 97.4 & \large \textbf{98.5} & \large 98.1  & \large 95.0 & - & \large 90.3 & \large 93.2 & \large 98.0 & - & - & - \\
        \large Leather    & \large \textbf{100}  & \large \textbf{100}  & \large \textbf{100}   & \large 97.2 & - & \large 95.1 & \large 96.5 & \large \textbf{100}  & - & - & - \\
        \large Meta Nut   & \large 77.0 & \large \textbf{92.5} & \large 66.1  & \large 60.2 & - & \large 58.2 & \large 63.4 & \large 65.6 & - & - & - \\
        \large Pill       & \large 81.0 & \large 81.2 & \large 66.3  & \large 59.7 & - & \large 57.3 & \large 62.4 & \large 65.1 & - & - & - \\
        \large Screw      & \large 57.4 & \large \textbf{57.9} & \large 55.9  & \large 49.6 & - & \large 51.2 & \large 53.5 & \large 54.8 & - & - & - \\
        \large Tile       & \large \textbf{99.9} & \large 99.2 & \large 99.8  & \large 89.5 & - & \large 90.2 & \large 88.7 & \large 99.5 & - & - & - \\
        \large Toothbrush & \large 84.4 & \large 85.2 & \large \textbf{86.7}  & \large 78.5 & - & \large 75.2 & \large 77.8 & \large 85.8 & - & - & - \\
        \large Transistor & \large 94.5 & \large \textbf{96.2} & \large 71.5  & \large 50.5 & - & \large 83.2 & \large 77.5 & \large 70.5 & - & - & - \\
        \large Wood       & \large 97.0 & \large 97.3 & \large 98.1  & \large 49.5 & - & \large 95.4 & \large 93.5 & \large 98.9 & - & - & - \\
        \large Zipper     & \large 97.4 & \large \textbf{97.5} & \large 50.3  & \large 48.5 & - & \large 45.2 & \large 48.6 & \large 51.2 & - & - & - \\
        \large Average    & \large 87.4 & \large \textbf{89.9} & \large 78.75 & \large 69.83 & \large 76.1  & \large 69.70 & \large 74.4 & \large 78.5 & \large 83.40 & \large 84.10 & \large 82.4 \\
    \hline
    \end{tabular}
}
\label{tab:mvtecv-1shot-imgroc}
\end{table}

% \begin{figure}
%     \centering
%     \includegraphics[width=1\linewidth]{iclr2023/fig/oneshot_pixelroc_mvtec_s_0.01.pdf}
%     \caption{Setting: New Few-shot Setting, K (number of shot)=1, Dataset: MVTec, Sampling Ratio: 0.01, Metrics: Pixel AUROC. RegAD$^*$2 shot }
%     \label{fig:oneshot_mvtec_img_auroc}
% \end{figure}

\begin{table}[ht]
% \setlength{\abovecaptionskip}{2pt}
%\caption{Results of anomaly detection. (MVTec, K=1 Shot, AUROC)}
\caption{Setting: Our Few-shot Setting, K (number of shot)=1, Dataset: MVTec, Sampling Ratio: 0.01, Metrics: Pixel AUROC. The number of shot for RegAD is 2. The data for PaDiM and PatchCore-10, PatchCore-25 are from \cite{roth2022towards}.}
% \normalsize
\renewcommand{\arraystretch}{1.2}
% \large
\resizebox{\textwidth}{!}{
    \begin{tabular}{l|ll lll llllllll}
    \hline
     \textbf{Category} & \textbf{Aug.(R)} & \textbf{GraphCore}  & \textbf{CFA} & \textbf{SPADE} & \textbf{PaDiM} & \textbf{STPM} & \textbf{RD4AD} & \textbf{PatchCore-1}& \textbf{PatchCore-10}& \textbf{PatchCore-25} & \textbf{RegAD} \\
        %  & \textbf{+PatchCore}
        %  \cite{rotation_augmentation} & &~\cite{Lee2022CFACF} &~\cite{cohen2020sub} &~\cite{defard2021padim} &~\cite{sheynin2021hierarchical} &~\cite{Wang2021StudentTeacherFP} &~\cite{Deng2022AnomalyDV} & ~\cite{roth2022towards}& & &~\cite{huang2022registration} \\
     \hline
        \large Bottle     & \large 98.5 & \large \textbf{99.8} & \large 93.2  & \large 85.2 & - & \large 84.3 & \large 81.7 & \large 93.0 & - & - & - \\
        \large Cable      & \large 95.1 & \large \textbf{96.2} & \large 88.2  & \large 78.2 & - & \large 50.9 & \large 64.8 & \large 87.0 & - & - & - \\
        \large Capsule    & \large 97.7 & \large \textbf{98.1} & \large 85.6  & \large 79.2 & - & \large 49.2 & \large 77.9 & \large 87.7 & - & - & - \\
        \large Carpet     & \large 99.1 & \large \textbf{99.3} & \large 97.5  & \large 95.2 & - & \large 60.5 & \large 72.5 & \large 98.8 & - & - & - \\
        \large Grid       & \large 70.5 & \large 76.9 & \large 81.3  & \large 75.6 & - & \large 61.2 & \large 74.3 & \large \textbf{84.1} & - & - & - \\
        \large Hazelnut   & \large 97.1 & \large \textbf{98.5} & \large 98.1  & \large 88.2 & - & \large 73.3 & \large 63.2 & \large 97.5 & - & - & - \\
        \large Leather    & \large 99.3 & \large \textbf{99.5} & \large 99.2  & \large 88.3 & - & \large 75.1 & \large 86.5 & \large 99.2 & - & - & - \\
        \large Meta Nut   & \large 93.2 & \large 92.5 & \large 89.5  & \large 58.5 & - & \large 51.1 & \large 68.7 & \large 90.1 & - & - & - \\
        \large Pill       & \large 95.7 & \large \textbf{96.2} & \large 91.2  & \large 54.2 & - & \large 49.3 & \large 65.6 & \large 90.4 & - & - & - \\
        \large Screw      & \large 92.0 & \large 93.4 & \large 96.5  & \large 69.6 & - & \large 51.2 & \large 59.7 & \large \textbf{95.8} & - & - & - \\
        \large Tile       & \large 95.8 & \large \textbf{96.8} & \large 81.5  & \large 81.5 & - & \large 57.2 & \large 88.7 & \large 82.7 & - & - & - \\
        \large Toothbrush & \large 97.9 & \large \textbf{98.5} & \large 93.8  & \large 75.5 & - & \large 65.2 & \large 77.8 & \large 93.0 & - & - & - \\
        \large Transistor & \large 93.6 & \large \textbf{96.2} & \large 79.5  & \large 73.5 & - & \large 43.2 & \large 77.5 & \large 78.8 & - & - & - \\
        \large Wood       & \large 93.1 & \large \textbf{94.3} & \large 91.8  & \large 89.5 & - & \large 45.4 & \large 93.5 & \large 90.7 & - & - & - \\
        \large Zipper     & \large \textbf{98.5} & \large 97.5 & \large 93.2  & \large 93.5 & - & \large 55.2 & \large 48.6 & \large 94.0 & - & - & - \\
        \large Average    & \large 94.47 & \large \textbf{95.60} & \large 90.67  & \large 79.07 & \large 88.20  & \large 58.15 & \large 69.03 & \large 90.85 & \large 92.00 & \large 92.40 & - \\
    \hline
    \end{tabular}
}
\label{tab:mvtecv-1shot-pixelroc}
\end{table}

% \begin{figure}
%     \centering
%     \includegraphics[width=1\linewidth]{iclr2023/fig/mvtec_2shot_pixelroc_s_0.01.png}
%     \caption{Setting: New Few-shot Setting, K (number of shot)=2, Dataset: MVTec, Sampling Ratio: 0.01, Metrics: Pixel AUROC. RegAD$^*$2 shot }
%     \label{fig:twoshot_mvtec_pixel_auroc}
% \end{figure}

\begin{table}[ht]
% \setlength{\abovecaptionskip}{2pt}
%\caption{Results of anomaly detection. (MVTec, K=1 Shot, AUROC)}
\caption{Setting: Our Few-shot Setting, K (number of shot)=2, Dataset: MVTec, Sampling Ratio: 0.01, Metrics: Image AUROC. The data for PaDiM and PatchCore-10, PatchCore-25 are from \cite{roth2022towards}.}
% \normalsize
\renewcommand{\arraystretch}{1.2}
% \large
\resizebox{\textwidth}{!}{
    \begin{tabular}{l|ll lll llllll}
    \hline
     \textbf{Category} & \textbf{Aug.(R)} & \textbf{GraphCore}  & \textbf{CFA} & \textbf{SPADE} & \textbf{PaDiM} & \textbf{STPM} & \textbf{RD4AD}  & \textbf{PatchCore-1}& \textbf{PatchCore-10}& \textbf{PatchCore-25} & \textbf{RegAD} \\
        %  & \textbf{+PatchCore} & &~\cite{Lee2022CFACF} &~\cite{cohen2020sub} &~\cite{Wang2021StudentTeacherFP} &~\cite{Deng2022AnomalyDV} & ~\cite{roth2022towards}&~\cite{huang2022registration}  \\
     \hline
        \large Bottle     & \large 99.7 & \large \textbf{99.8} & \large 93.7  & \large 95.7 & - & \large 93.8 & \large 91.2 & \large 99.7 & - & - & \large 99.4 \\
        \large Cable      & \large 94.7 & \large \textbf{95.2} & \large 89.3  & \large 60.4 & - & \large 60.2 & \large 65.3 & \large 94.9 & - & - & \large 65.1 \\
        \large Capsule    & \large 66.5 & \large \textbf{73.2} & \large 53.4  & \large 48.7 & - & \large 45.2 & \large 50.5 & \large 67.2 & - & - & \large 67.5 \\
        \large Carpet     & \large \textbf{99.4} & \large \textbf{99.4} & \large 97.6  & \large 92.1 & - & \large 90.8 & \large 92.8 & \large 99.1 & - & - & \large 96.5 \\
        \large Grid       & \large 75.7 & \large 81.5 & \large 80.4  & \large 75.8 & - & \large 72.6 & \large 75.2 & \large 61.7 & - & - & \large \textbf{84.0} \\
        \large Hazelnut   & \large \textbf{99.7} & \large 99.5 & \large 99.4  & \large 95.2 & - & \large 90.3 & \large 93.4 & \large 93.5 & - & - & \large 96.0 \\
        \large Leather    & \large \textbf{100}  & \large \textbf{100}  & \large \textbf{100}   & \large 97.9 & - & \large 95.8 & \large 96.7 & \large \textbf{100}  & - & - & \large 99.4 \\
        \large Meta Nut   & \large 95.0 & \large \textbf{96.3} & \large 68.6  & \large 61.3 & - & \large 59.4 & \large 63.4 & \large 92.0 & - & - & \large 91.4 \\
        \large Pill       & \large 87.8 & \large \textbf{88.6} & \large 67.4  & \large 60.2 & - & \large 58.7 & \large 62.8 & \large 87.4 & - & - & \large 81.3 \\
        \large Screw      & \large 63.6 & \large \textbf{65.7} & \large 58.2  & \large 51.3 & - & \large 51.9 & \large 54.3 & \large 48.3 & - & - & \large 52.5 \\
        \large Tile       & \large \textbf{100}  & \large \textbf{100}  & \large 99.8  & \large 90.2 & - & \large 91.4 & \large 88.9 & \large \textbf{100}  & - & - & \large 94.3 \\
        \large Toothbrush & \large 83.6 & \large \textbf{87.3} & \large 86.9  & \large 80.2 & - & \large 76.5 & \large 77.1 & \large 83.9 & - & - & \large 86.6 \\
        \large Transistor & \large 96.3 & \large \textbf{97.1} & \large 72.5  & \large 51.6 & - & \large 82.4 & \large 78.1 & \large 96.9 & - & - & \large 86.0 \\
        \large Wood       & \large 97.1 & \large 97.5 & \large 98.2  & \large 50.3 & - & \large 95.8 & \large 93.7 & \large 97.2 & - & - & \large \textbf{99.2} \\
        \large Zipper     & \large 96.9 & \large \textbf{97.5} & \large 50.5  & \large 49.4 & - & \large 47.6 & \large 49.5 & \large 95.3 & - & - & \large 86.3 \\
        \large Average    & \large 90.40 & \large \textbf{91.91} & \large 81.06 & \large 70.69 & \large 78.90 & \large 74.16 & \large 75.53 & \large 87.81 & \large 86.40 & \large 87.20 & \large 85.70 \\
    \hline
    \end{tabular}
}
\label{tab:mvtecv-2shot-imgroc}
\end{table}

\begin{table}[ht]
% \setlength{\abovecaptionskip}{2pt}
%\caption{Results of anomaly detection. (MVTec, K=1 Shot, AUROC)}
\caption{Setting: Ours Few-shot Setting, K (number of shot)=2, Dataset: MVTec, Sampling Ratio: 0.01, Metrics: Pixel AUROC. The data for PaDiM and PatchCore-10, PatchCore-25 are from \cite{roth2022towards}.}
% \normalsize
\renewcommand{\arraystretch}{1.2}
% \large
\resizebox{\textwidth}{!}{
    \begin{tabular}{l|ll lll lllllll}
    \hline
     \textbf{Category} & \textbf{Aug.(R)} & \textbf{GraphCore}  & \textbf{CFA} & \textbf{SPADE} & \textbf{PaDiM} & \textbf{STPM} & \textbf{RD4AD} & \textbf{PatchCore-1}& \textbf{PatchCore-10}& \textbf{PatchCore-25} & \textbf{RegAD} \\
        %  & \textbf{+PatchCore}
        %  \cite{rotation_augmentation} & &~\cite{Lee2022CFACF} &~\cite{cohen2020sub} &~\cite{defard2021padim} &~\cite{sheynin2021hierarchical} &~\cite{Wang2021StudentTeacherFP} &~\cite{Deng2022AnomalyDV} & ~\cite{roth2022towards}& & &~\cite{huang2022registration} \\
     \hline
        \large Bottle     & \large 98.6 & \large 	\textbf{99.8} & \large 	93.5 & \large 	86.8 & - & \large 	84.6 & \large 	81.7 & \large 	98.5 & - & - & \large 98.0 \\
        \large Cable      & \large 96.2 & \large 	96.3 & \large 	88.9 & \large 	78.6 & - & \large 	51.6 & \large 	65.4 & \large 	\textbf{97.8} & - & - & \large 91.7 \\
        \large Capsule    & \large 97.7 & \large 	\textbf{97.8} & \large 	85.9 & \large 	79.8 & - & \large 	59.2 & \large 	78.2 & \large 	97.7 & - & - & \large 97.3 \\
        \large Carpet     & \large 99.1 & \large 	\textbf{99.6} & \large 	97.9 & \large 	95.6 & - & \large 	60.5 & \large 	74.2 & \large 	99.0 & - & - & \large 98.9 \\
        \large Grid       & \large 79.8 & \large 	80.6 & \large 	\textbf{81.4} & \large 	75.9 & - & \large 	61.2 & \large 	76.3 & \large 	67.5 & - & - & \large 77.4 \\
        \large Hazelnut   & \large 97.9 & \large 	\textbf{98.2} & \large 	98.2 & \large 	88.9 & - & \large 	74.5 & \large 	64.8 & \large 	96.4 & - & - & \large 98.1 \\
        \large Leather    & \large 99.3 & \large 	\textbf{99.4} & \large 	99.3 & \large 	89.2 & - & \large 	75.2 & \large 	86.5 & \large 	99.3 & - & - & \large 98.0 \\
        \large Meta Nut   & \large 96.8 & \large 	\textbf{98.1} & \large 	89.7 & \large 	59.5 & - & \large 	51.1 & \large 	68.9 & \large 	97.1 & - & - & \large 96.9 \\
        \large Pill       & \large 93.9 & \large 	94.1 & \large 	91.5 & \large 	57.2 & - & \large 	49.9 & \large 	70.2 & \large 	\textbf{96.8} & - & - & \large 93.6 \\
        \large Screw      & \large 96.0 & \large 	96.5 & \large 	\textbf{96.7} & \large 	70.2 & - & \large 	51.8 & \large 	60.8 & \large 	90.8 & - & - & \large 94.4 \\
        \large Tile       & \large 99.3 & \large 	\textbf{96.8} & \large 	81.8 & \large 	82.3 & - & \large 	58.2 & \large 	59.2 & \large 	96.0 & - & - & \large 94.3 \\
        \large Toothbrush & \large 98.2 & \large 	\textbf{98.6} & \large 	93.9 & \large 	76.8 & - & \large 	66.3 & \large 	78.3 & \large 	98.2 & - & - & \large 98.2 \\
        \large Transistor & \large 94.1 & \large 	\textbf{99.2} & \large 	80.3 & \large 	73.6 & - & \large 	47.5 & \large 	67.8 & \large 	95.0 & - & - & \large 93.4 \\
        \large Wood       & \large 98.4 & \large 	\textbf{99.5} & \large 	92.4 & \large 	89.7 & - & \large 	48.4 & \large 	93.8 & \large 	93.0 & - & - & \large 93.5 \\
        \large Zipper     & \large 99.0 & \large 	\textbf{99.3} & \large 	94.1 & \large 	93.7 & - & \large 	56.3 & \large 	51.2 & \large 	98.2 & - & - & \large 95.1 \\
        \large Average    & \large 96.29 & \large 	\textbf{96.92} & \large 	91.03 & \large 	79.85 & 90.50 & \large 	59.75 & \large 	71.82 & \large 	94.75 & 93.10 & 93.30 & \large 94.59 \\
    \hline
    \end{tabular}
}
\label{tab:mvtecv-2shot-pixelroc}
\end{table}

% \begin{figure}
%     \centering
%     \includegraphics[width=1\linewidth]{iclr2023/fig/mvtec_4shot_pixelroc_s_0.01.jpeg}
%     \caption{Setting: New Few-shot Setting, K (number of shot)=4, Dataset: MVTec, Sampling Ratio: 0.01, Metrics: Pixel AUROC}
%     \label{fig:oneshot_mvtec_img_auroc}
% \end{figure}

\begin{table}[ht]
% \setlength{\abovecaptionskip}{2pt}
%\caption{Results of anomaly detection. (MVTec, K=1 Shot, AUROC)}
\caption{Setting: New Few-shot Setting, K (number of shot)=4, Dataset: MVTec, Sampling Ratio: 0.01, Metrics: Image AUROC}
% \normalsize
\renewcommand{\arraystretch}{1.2}
% \large
\resizebox{\textwidth}{!}{
    \begin{tabular}{l|ll lll lll}
    \hline
     \textbf{Category} & \textbf{Aug.(R)} & \textbf{GraphCore}  & \textbf{CFA} & \textbf{SPADE} & \textbf{STPM} & \textbf{RD4AD}  & \textbf{PatchCore} & \textbf{RegAD} \\
        %  & \textbf{+PatchCore} & &~\cite{Lee2022CFACF} &~\cite{cohen2020sub} &~\cite{Wang2021StudentTeacherFP} &~\cite{Deng2022AnomalyDV} & ~\cite{roth2022towards}&~\cite{huang2022registration}  \\
     \hline
        \large Bottle     & \large 99.7 & \large \textbf{99.8} & \large 94.2  & \large 95.8  & \large 93.9 & \large 92.1 & \large 99.6 &  \large 99.4 \\
        \large Cable      & \large 94.1 & \large 95.2 & \large 91.2  & \large 61.3  & \large 61.3 & \large 68.4 & \large \textbf{97.4} &  \large 76.1 \\
        \large Capsule    & \large 66.2 & \large \textbf{74.5} & \large 56.2  & \large 48.7  & \large 47.4 & \large 51.7 & \large 66.3 &  \large 72.4 \\
        \large Carpet     & \large \textbf{99.6} & \large 99.4 & \large 97.6  & \large 92.5  & \large 91.5 & \large 93.2 & \large 99.0 &  \large 97.9 \\
        \large Grid       & \large 77.9 & \large 81.6 & \large 81.5  & \large 76.2  & \large 75.3 & \large 76.4 & \large 63.0 &  \large \textbf{91.2} \\
        \large Hazelnut   & \large \textbf{99.9} & \large 99.5 & \large 99.4  & \large 95.6  & \large 91.4 & \large 93.8 & \large 92.8 &  \large 95.8 \\
        \large Leather    & \large \textbf{100} & \large \textbf{100}  & \large \textbf{100}   & \large 98.2  & \large 96.9 & \large 96.8 & \large \textbf{100}  &  \large \textbf{100} \\
        \large Meta Nut   & \large 95.9 & \large \textbf{96.2} & \large 91.3  & \large 62.5  & \large 60.8 & \large 65.3 & \large 94.7 &  \large 94.6 \\
        \large Pill       & \large \textbf{89.3} & \large 88.2 & \large 85.6  & \large 61.8  & \large 61.3 & \large 62.8 & \large 89.0 &  \large 80.8 \\
        \large Screw      & \large 63.9 & \large \textbf{68.9} & \large 49.2  & \large 52.9  & \large 52.8 & \large 55.7 & \large 54.1 &  \large 56.6 \\
        \large Tile       & \large \textbf{100}  & \large \textbf{100} & \large 99.8  & \large 91.3  & \large 90.4 & \large 90.8 & \large 100  &  \large 95.5 \\
        \large Toothbrush & \large 94.4 & \large \textbf{95.2} & \large 87.2  & \large 81.7  & \large 80.4 & \large 76.7 & \large \textbf{95.2} &  \large 90.9 \\
        \large Transistor & \large 98.5 & \large \textbf{99.2} & \large 95.8  & \large 52.5  & \large 82.4 & \large 79.3 & \large 98.4 &  \large 85.2 \\
        \large Wood       & \large 97.4 & \large 97.9 & \large \textbf{98.6}  & \large 51.4  & \large 95.8 & \large 94.2 & \large 97.4 &  \large \textbf{98.6} \\
        \large Zipper     & \large 96.9 & \large \textbf{98.2} & \large 94.3  & \large 52.2  & \large 47.6 & \large 56.7 & \large 95.5 &  \large 88.5 \\
        \large Average    & \large 92.22 & \large \textbf{92.92} & \large 84.97 & \large 71.64  & \large 74.77 & \large 76.93 & \large 89.49  & \large 88.23 \\
    \hline
    \end{tabular}
}
\label{tab:mvtecv-4shot-imgroc}
\end{table}

\begin{table}[ht]
% \setlength{\abovecaptionskip}{2pt}
%\caption{Results of anomaly detection. (MVTec, K=1 Shot, AUROC)}
\caption{Setting: New Few-shot Setting, K (number of shot)=4, Dataset: MVTec, Sampling Ratio: 0.01, Metrics: Pixel AUROC}
% \normalsize
\renewcommand{\arraystretch}{1.2}
% \large
\resizebox{\textwidth}{!}{
    \begin{tabular}{l|ll lll lll}
    \hline
     \textbf{Category} & \textbf{Aug.(R)} & \textbf{GraphCore}  & \textbf{CFA} & \textbf{SPADE} & \textbf{STPM} & \textbf{RD4AD}  & \textbf{PatchCore} & \textbf{RegAD} \\
        %  & \textbf{+PatchCore} & &~\cite{Lee2022CFACF} &~\cite{cohen2020sub} &~\cite{Wang2021StudentTeacherFP} &~\cite{Deng2022AnomalyDV} & ~\cite{roth2022towards}&~\cite{huang2022registration}  \\
     \hline
        \large Bottle     & \large 98.6  & \large \textbf{99.8}  & \large 93.6  & \large 86.9  & \large 84.9  & \large 81.8  & \large 98.6  & \large 98.4 \\
        \large Cable      & \large 96.6  & \large 96.9  & \large 89.1  & \large 78.7  & \large 52.2  & \large 66.2  & \large \textbf{97.9}  & \large 92.7 \\
        \large Capsule    & \large 97.7  & \large \textbf{97.9}  & \large 86.2  & \large 80.1  & \large 59.3  & \large 78.4  & \large 97.7  & \large 97.6 \\
        \large Carpet     & \large 99.1  & \large \textbf{99.6}  & \large 98.2  & \large 95.0  & \large 60.6  & \large 74.8  & \large 99.0  & \large 98.9 \\
        \large Grid       & \large 81.9  & \large 82.3  & \large 82.5  & \large 76.1  & \large 61.8  & \large 76.9  & \large 70.6  & \large \textbf{85.7} \\
        \large Hazelnut   & \large 98.3  & \large \textbf{99.1}  & \large 98.5  & \large 89.1  & \large 74.9  & \large 65.2  & \large 97.0  & \large 98.0 \\
        \large Leather    & \large 99.3  & \large \textbf{99.6}  & \large 99.3  & \large 89.3  & \large 75.3  & \large 86.7  & \large 96.9  & \large 99.1 \\
        \large Meta Nut   & \large 96.8  & \large \textbf{98.1}  & \large 89.9  & \large 60.2  & \large 51.8  & \large 69.2  & \large 97.0  & \large 97.8 \\
        \large Pill       & \large 97.0  & \large \textbf{97.5}  & \large 91.6  & \large 58.2  & \large 50.6  & \large 70.4  & \large 96.9  & \large 97.4 \\
        \large Screw      & \large 93.8  & \large \textbf{96.5}  & \large 96.8  & \large 71.3  & \large 51.9  & \large 60.9  & \large 92.1  & \large 95.0 \\
        \large Tile       & \large 95.7  & \large \textbf{96.7}  & \large 82.3  & \large 82.4  & \large 58.5  & \large 59.5  & \large 96.0  & \large 94.9 \\
        \large Toothbrush & \large 98.8  & \large \textbf{98.9}  & \large 94.2  & \large 76.9  & \large 66.9  & \large 78.9  & \large 98.8  & \large 98.5 \\
        \large Transistor & \large 94.1  & \large \textbf{99.3}  & \large 80.5  & \large 74.2  & \large 57.5  & \large 67.9  & \large 95.0  & \large 93.8 \\
        \large Wood       & \large 93.2  & \large \textbf{99.5}  & \large 92.6  & \large 90.4  & \large 48.9  & \large 94.2  & \large 93.1  & \large 94.7 \\
        \large Zipper     & \large 98.4  & \large \textbf{99.3}  & \large 94.8  & \large 93.8  & \large 56.4  & \large 52.3  & \large 98.3  & \large 94.0 \\
        \large Average    & \large 95.95  & \large \textbf{97.40}  & \large 91.34  & \large 80.17  & \large 60.77  & \large 72.22  & \large 94.99  & \large 95.77 \\
    \hline
    \end{tabular}
}
\label{tab:mvtecv-4shot-pixelroc}
\end{table}

\begin{table}[ht]
% \setlength{\abovecaptionskip}{2pt}
%\caption{Results of anomaly detection. (MVTec, K=1 Shot, AUROC)}
\caption{Setting: New Few-shot Setting, K (number of shot)=8, Dataset: MVTec, Sampling Ratio: 0.01, Metrics: Image AUROC}
% \normalsize
\renewcommand{\arraystretch}{1.2}
% \large
\resizebox{\textwidth}{!}{
    \begin{tabular}{l|ll lll lll}
    \hline
     \textbf{Category} & \textbf{Aug.(R)} & \textbf{GraphCore}  & \textbf{CFA} & \textbf{SPADE} & \textbf{STPM} & \textbf{RD4AD}  & \textbf{PatchCore} & \textbf{RegAD} \\
        %  & \textbf{+PatchCore} & &~\cite{Lee2022CFACF} &~\cite{cohen2020sub} &~\cite{Wang2021StudentTeacherFP} &~\cite{Deng2022AnomalyDV} & ~\cite{roth2022towards}&~\cite{huang2022registration}  \\
     \hline
        \large Bottle     & \large \textbf{100}  & \large 99.8 & \large 95.1 & \large 95.9 & \large 94.1 & \large 92.8 & \large 99.6 & \large 99.8 \\
        \large Cable      & \large 94.1 & \large 95.2 & \large 91.8 & \large 63.5 & \large 62.6 & \large 69.2 & \large \textbf{97.4} & \large 80.6 \\
        \large Capsule    & \large 89.7 & \large \textbf{90.5} & \large 69.5 & \large 58.9 & \large 57.8 & \large 58.5 & \large 85.3 & \large 76.3 \\
        \large Carpet     & \large 98.5 & \large \textbf{99.5} & \large 97.6 & \large 92.7 & \large 91.6 & \large 93.8 & \large 99.0 & \large 98.5 \\
        \large Grid       & \large \textbf{92.7} & \large 92.3 & \large 85.6 & \large 77.3 & \large 76.9 & \large 77.9 & \large 83.1 & \large 91.5 \\
        \large Hazelnut   & \large \textbf{100}  & \large \textbf{100}  & \large 99.4 & \large 96.5 & \large 91.8 & \large 94.2 & \large 99.8 & \large 96.5 \\
        \large Leather    & \large \textbf{100}  & \large \textbf{100}  & \large \textbf{100}  & \large 98.7 & \large 97.2 & \large 97.2 & \large \textbf{100}  & \large \textbf{100}  \\ 
        \large Meta Nut   & \large 96.8 & \large 97.9 & \large 92.3 & \large 68.9 & \large 61.3 & \large 65.6 & \large 95.1 & \large \textbf{98.3} \\
        \large Pill       & \large 90.1 & \large \textbf{91.1} & \large 88.9 & \large 63.9 & \large 64.2 & \large 63.6 & \large 89.6 & \large 80.6 \\
        \large Screw      & \large 79.4 & \large 80.1 & \large 65.4 & \large 56.4 & \large 55.9 & \large 59.3 & \large 74.1 & \large 63.4 \\
        \large Tile       & \large 99.3 & \large \textbf{100}  & \large 99.8 & \large 91.8 & \large 91.2 & \large 91.2 & \large \textbf{100}  & \large 97.4 \\
        \large Toothbrush & \large 94.6 & \large 95.1 & \large 88.9 & \large 82.9 & \large 82.3 & \large 77.9 & \large 96.8 & \large \textbf{98.5} \\
        \large Transistor & \large 98.2 & \large \textbf{99.2} & \large 96.2 & \large 58.9 & \large 84.6 & \large 81.2 & \large 98.9 & \large 93.4 \\
        \large Wood       & \large 98.7 & \large 98.9 & \large 98.9 & \large 61.3 & \large 95.8 & \large 95.6 & \large 97.5 & \large \textbf{99.4} \\
        \large Zipper     & \large 99.0 & \large \textbf{99.2} & \large 94.5 & \large 62.5 & \large 57.2 & \large 58.9 & \large 98.4 & \large 94.0 \\
        \large Average    & \large 95.41 & \large \textbf{95.92} & \large 90.93 & \large 75.34 & \large 77.63 & \large 78.46 & \large 94.31 & \large 91.21 \\
    \hline
    \end{tabular}
}
\label{tab:mvtecv-8shot-imgroc}
\end{table}

% \begin{figure}
%     \centering
%     \includegraphics[width=1\linewidth]{iclr2023/fig/mvtec_fourshot_imgroc_s_0.01.png}
%     \caption{Setting: New Few-shot Setting, K (number of shot)=4, Dataset: MVTec, Sampling Ratio: 0.01, Metrics: Image AUROC}
%     \label{fig:fourshot_mvtec_img_auroc}
% \end{figure}

\begin{table}[ht]
% \setlength{\abovecaptionskip}{2pt}
%\caption{Results of anomaly detection. (MVTec, K=1 Shot, AUROC)}
\caption{Setting: New Few-shot Setting, K (number of shot)=8, Dataset: MVTec, Sampling Ratio: 0.01, Metrics: Pixel AUROC}
% \normalsize
\renewcommand{\arraystretch}{1.2}
% \large
\resizebox{\textwidth}{!}{
    \begin{tabular}{l|ll lll lll}
    \hline
     \textbf{Category} & \textbf{Aug.(R)} & \textbf{GraphCore}  & \textbf{CFA} & \textbf{SPADE} & \textbf{STPM} & \textbf{RD4AD}  & \textbf{PatchCore} & \textbf{RegAD} \\
        %  & \textbf{+PatchCore} & &~\cite{Lee2022CFACF} &~\cite{cohen2020sub} &~\cite{Wang2021StudentTeacherFP} &~\cite{Deng2022AnomalyDV} & ~\cite{roth2022towards}&~\cite{huang2022registration}  \\
     \hline
        \large Bottle     & \large 98.6  & \large \textbf{99.8}  & \large 93.6  & \large 87.1  & \large 85.2  & \large 82.1  & \large 98.7  & \large 97.5 \\
        \large Cable      & \large 97.0  & \large 97.2  & \large 89.2  & \large 78.9  & \large 53.3  & \large 68.2  & \large \textbf{98.3}  & \large 94.9 \\
        \large Capsule    & \large 98.3  & \large \textbf{98.5}  & \large 86.5  & \large 80.2  & \large 59.3  & \large 78.5  & \large 98.4  & \large 98.2 \\
        \large Carpet     & \large 99.1  & \large \textbf{99.7}  & \large 98.4  & \large 95.1  & \large 60.7  & \large 79.2  & \large 99.2  & \large 98.9 \\
        \large Grid       & \large 82.5  & \large 83.7  & \large 82.8  & \large 77.2  & \large 61.8  & \large 76.9  & \large 71.5  & \large \textbf{88.7} \\
        \large Hazelnut   & \large 98.4  & \large \textbf{99.2}  & \large 98.6  & \large 89.5  & \large 74.9  & \large 65.5  & \large 97.2  & \large 98.5 \\
        \large Leather    & \large 99.4  & \large \textbf{99.6}  & \large 99.4  & \large 90.2  & \large 75.3  & \large 86.9  & \large 99.4  & \large 98.9 \\
        \large Meta Nut   & \large 97.3  & \large \textbf{98.9}  & \large 89.9  & \large 60.5  & \large 54.6  & \large 69.5  & \large 97.5  & \large 96.9 \\
        \large Pill       & \large 98.1  & \large \textbf{98.4}  & \large 91.7  & \large 58.2  & \large 55.7  & \large 70.5  & \large 98.1  & \large 97.8 \\
        \large Screw      & \large 94.2  & \large 96.6  & \large 96.9  & \large 71.4  & \large 52.3  & \large 61.9  & \large 92.5  & \large \textbf{97.1} \\
        \large Tile       & \large 96.8  & \large \textbf{97.4}  & \large 83.4  & \large 82.5  & \large 58.9  & \large 60.8  & \large 96.3  & \large 95.2 \\
        \large Toothbrush & \large \textbf{99.2}  & \large \textbf{99.2}  & \large 94.5  & \large 77.2  & \large 66.9  & \large 79.1  & \large \textbf{99.2}  & \large 98.7 \\
        \large Transistor & \large 95.2  & \large \textbf{99.4}  & \large 81.5  & \large 74.5  & \large 58.2  & \large 67.9  & \large 95.7  & \large 96.8 \\
        \large Wood       & \large 93.8  & \large \textbf{99.7}  & \large 92.7  & \large 90.4  & \large 49.2  & \large 94.5  & \large 93.4  & \large 94.6 \\
        \large Zipper     & \large 98.6  & \large \textbf{99.7}  & \large 94.9  & \large 94.2  & \large 57.8  & \large 52.8  & \large 98.6  & \large 97.4 \\
        \large Average    & \large 96.43  & \large \textbf{97.80}  & \large 91.60  & \large 80.47  & \large 61.61  & \large 72.95  & \large 95.60  & \large 96.67 \\
    \hline
    \end{tabular}
}
\label{tab:mvtecv-8shot-pixelroc}
\end{table}

\begin{table}[ht]
% \setlength{\abovecaptionskip}{2pt}
%\caption{Results of anomaly detection. (MVTec, K=1 Shot, AUROC)}
\caption{Setting: New Few-shot Setting, K (number of shot)=1, Dataset: MPDD, Sampling Ratio: 0.01, Metrics: Image AUROC}
% \normalsize
\renewcommand{\arraystretch}{1.2}
% \large
\resizebox{\textwidth}{!}{
    \begin{tabular}{l|ll lll lll}
    \hline
     \textbf{Category} & \textbf{Aug.(R)} & \textbf{GraphCore}  & \textbf{CFA} & \textbf{SPADE} & \textbf{STPM} & \textbf{RD4AD}  & \textbf{PatchCore} & \textbf{RegAD} \\
        %  & \textbf{+PatchCore} & &~\cite{Lee2022CFACF} &~\cite{cohen2020sub} &~\cite{Wang2021StudentTeacherFP} &~\cite{Deng2022AnomalyDV} & ~\cite{roth2022towards}&~\cite{huang2022registration}  \\
     \hline
        \large Bracket Black & \large 64.8 & \large 65.9 & \large 64.1  & \large 62.1  & \large \textbf{93.2} & \large 91.2 & \large 58.2 &  \large - \\
        \large Bracket Brown & \large 75.0 & \large \textbf{76.8} & \large 65.4  & \large 59.2  & \large 59.8 & \large 58.3 & \large 70.6 &  \large - \\
        \large Bracket White & \large 88.6 & \large \textbf{89.2} & \large 68.2  & \large 68.2  & \large 43.2 & \large 44.7 & \large 69.3 &  \large - \\
        \large Connector     & \large 98.3 & \large \textbf{98.7} & \large 58.5  & \large 58.5  & \large 90.5 & \large 92.5 & \large 59.0 &  \large - \\
        \large Metal Plate   & \large \textbf{99.9} & \large \textbf{99.9} & \large 62.1  & \large 63.2  & \large 71.2 & \large 74.3 & \large 64.1 &  \large - \\
        \large Tubes         & \large 76.6 & \large \textbf{77.8} & \large 34.2  & \large 33.8  & \large 65.1 & \large 44.2 & \large 34.1 &  \large - \\
        \large Average       & \large 83.87 & \large \textbf{84.72} & \large 58.75 & \large 57.50  & \large 59.20 & \large 67.53 & \large 59.22  & \large 57.8 \\
    \hline
    \end{tabular}
}
\label{tab:mpdd-1shot-imgroc}
\end{table}

% \begin{figure}
%     \centering
%     \includegraphics[width=1\linewidth]{iclr2023/fig/mpdd_1shot_imgroc_s_0.01.png}
%     \caption{Setting: New Few-shot Setting, K (number of shot)=1, Dataset: MPDD, Sampling Ratio: 0.01, Metrics: Image AUROC}
%     \label{fig:oneshot_mpdd_img_auroc}
% \end{figure}

\begin{table}[ht]
\caption{Setting: New Few-shot Setting, K (number of shot)=1, Dataset: MPDD, Sampling Ratio: 0.01, Metrics: Pixel AUROC}
\renewcommand{\arraystretch}{1.2}
\resizebox{\textwidth}{!}{
    \begin{tabular}{l|ll lll lll}
    \hline
     \textbf{Category} & \textbf{Aug.(R)} & \textbf{GraphCore}  & \textbf{CFA} & \textbf{SPADE} & \textbf{STPM} & \textbf{RD4AD}  & \textbf{PatchCore} & \textbf{RegAD} \\
        %  & \textbf{+PatchCore} & &~\cite{Lee2022CFACF} &~\cite{cohen2020sub} &~\cite{Wang2021StudentTeacherFP} &~\cite{Deng2022AnomalyDV} & ~\cite{roth2022towards}&~\cite{huang2022registration}  \\
     \hline
        \large Bracket Black & \large 91.7  & \large \textbf{92.3}  & \large 75.2  & \large 72.4  & \large 74.5  & \large 72.5  & \large 78.8 & \large - \\
        \large Bracket Brown & \large 91.8  & \large \textbf{92.2}  & \large 77.2  & \large 71.8  & \large 72.9  & \large 72.3  & \large 76.8 & \large - \\
        \large Bracket White & \large 97.0  & \large \textbf{97.3}  & \large 69.8  & \large 65.4  & \large 63.1  & \large 61.3  & \large 67.8 & \large - \\
        \large Connector     & \large 97.0  & \large \textbf{97.5}  & \large 88.9  & \large 82.4  & \large 82.1  & \large 81.7  & \large 85.0 & \large - \\
        \large Metal Plate   & \large 98.1  & \large \textbf{98.9}  & \large 83.1  & \large 75.2  & \large 83.2  & \large 75.4  & \large 84.1 & \large - \\
        \large Tubes         & \large 92.4  & \large \textbf{92.8}  & \large 71.7  & \large 76.2  & \large 74.5  & \large 76.1  & \large 78.2 & \large - \\
        \large Average       & \large 94.67  & \large \textbf{95.17}  & \large 77.65  & \large 73.90  & \large 75.05  & \large 73.22  & \large 78.45 & \large - \\
    \hline
    \end{tabular}
}
\label{tab:mpdd-1shot-pixelroc}
\end{table}

\begin{table}[ht]
% \setlength{\abovecaptionskip}{2pt}
%\caption{Results of anomaly detection. (MVTec, K=1 Shot, AUROC)}
\caption{Setting: New Few-shot Setting, K (number of shot)=2, Dataset: MPDD, Sampling Ratio: 0.01, Metrics: Image AUROC}
% \normalsize
\renewcommand{\arraystretch}{1.2}
% \large
\resizebox{\textwidth}{!}{
    \begin{tabular}{l|ll lll lll}
    \hline
     \textbf{Category} & \textbf{Aug.(R)} & \textbf{GraphCore}  & \textbf{CFA} & \textbf{SPADE} & \textbf{STPM} & \textbf{RD4AD}  & \textbf{PatchCore} & \textbf{RegAD} \\
        %  & \textbf{+PatchCore} & &~\cite{Lee2022CFACF} &~\cite{cohen2020sub} &~\cite{Wang2021StudentTeacherFP} &~\cite{Deng2022AnomalyDV} & ~\cite{roth2022towards}&~\cite{huang2022registration}  \\
     \hline
        \large Bracket Black & \large 66.8 & \large 67.0 & \large 54.3  & \large 62.4  & \large \textbf{94.5} & \large 91.7 & \large 58.6 &  \large 63.3 \\
        \large Bracket Brown & \large 76.1 & \large \textbf{77.2} & \large 66.8  & \large 59.5  & \large 62.3 & \large 58.8 & \large 70.7 &  \large 59.4 \\
        \large Bracket White & \large 87.2 & \large \textbf{89.4} & \large 68.7  & \large 67.2  & \large 53.8 & \large 55.6 & \large 70.4 &  \large 55.6 \\
        \large Connector     & \large 98.6 & \large \textbf{98.9} & \large 58.5  & \large 59.2  & \large 51.6 & \large 53.7 & \large 59.2 &  \large 73.0 \\
        \large Metal Plate   & \large \textbf{99.9} & \large \textbf{99.9} & \large 62.7  & \large 64.2  & \large 62.4 & \large 65.2 & \large 64.1 &  \large 61.7 \\
        \large Tubes         & \large 79.2 & \large \textbf{79.8} & \large 40.7  & \large 35.6  & \large 49.6 & \large 45.9 & \large 34.3 &  \large 67.1 \\
        \large Average       & \large 84.63 & \large \textbf{85.37} & \large 58.62 & \large 58.02  & \large 62.37 & \large 61.82 & \large 59.55  & \large 63.35 \\
    \hline
    \end{tabular}
}
\label{tab:mpdd-2shot-imgroc}
\end{table}

\begin{table}[ht]
\caption{Setting: New Few-shot Setting, K (number of shot)=2, Dataset: MPDD, Sampling Ratio: 0.01, Metrics: Pixel AUROC}
\renewcommand{\arraystretch}{1.2}
\resizebox{\textwidth}{!}{
    \begin{tabular}{l|ll lll lll}
    \hline
     \textbf{Category} & \textbf{Aug.(R)} & \textbf{GraphCore}  & \textbf{CFA} & \textbf{SPADE} & \textbf{STPM} & \textbf{RD4AD}  & \textbf{PatchCore} & \textbf{RegAD} \\
        %  & \textbf{+PatchCore} & &~\cite{Lee2022CFACF} &~\cite{cohen2020sub} &~\cite{Wang2021StudentTeacherFP} &~\cite{Deng2022AnomalyDV} & ~\cite{roth2022towards}&~\cite{huang2022registration}  \\
     \hline
        \large Bracket Black & \large 92.1 & \large \textbf{92.5} & \large 75.8 & \large 72.8 & \large 75.1 & \large 75.4 & \large 78.9 & \large - \\
        \large Bracket Brown & \large 91.9 & \large \textbf{92.6} & \large 77.5 & \large 71.9 & \large 73.2 & \large 73.4 & \large 76.9 & \large - \\
        \large Bracket White & \large 97.1 & \large \textbf{97.5} & \large 70.8 & \large 72.4 & \large 64.2 & \large 62.4 & \large 68.1 & \large - \\
        \large Connector     & \large 97.2 & \large \textbf{97.7} & \large 88.2 & \large 82.8 & \large 83.4 & \large 82.3 & \large 85.2 & \large - \\
        \large Metal Plate   & \large 98.4 & \large \textbf{99.1} & \large 84.3 & \large 75.9 & \large 83.2 & \large 76.5 & \large 86.3 & \large - \\
        \large Tubes         & \large 92.6 & \large \textbf{93.1} & \large 72.8 & \large 76.8 & \large 75.6 & \large 77.1 & \large 79.5 & \large - \\
        \large Average       & \large 94.88 & \large \textbf{95.42} & \large 78.23 & \large 75.43 & \large 75.78 & \large 74.52 & \large 79.15 & \large 93.2 \\
    \hline
    \end{tabular}
}
\label{tab:mpdd-2shot-pixelroc}
\end{table}

% \begin{figure}
%     \centering
%     \includegraphics[width=1\linewidth]{iclr2023/fig/mpdd_2shot_imgroc_s_0.01.png}
%     \caption{Setting: New Few-shot Setting, K (number of shot)=2, Dataset: MPDD, Sampling Ratio: 0.01, Metrics: Image AUROC}
%     \label{fig:twoshot_mpdd_img_auroc}
% \end{figure}

\begin{table}[ht]
% \setlength{\abovecaptionskip}{2pt}
%\caption{Results of anomaly detection. (MVTec, K=1 Shot, AUROC)}
\caption{Setting: New Few-shot Setting, K (number of shot)=4, Dataset: MPDD, Sampling Ratio: 0.01, Metrics: Image AUROC}
% \normalsize
\renewcommand{\arraystretch}{1.2}
% \large
\resizebox{\textwidth}{!}{
    \begin{tabular}{l|ll lll lll}
    \hline
     \textbf{Category} & \textbf{Aug.(R)} & \textbf{GraphCore}  & \textbf{CFA} & \textbf{SPADE} & \textbf{STPM} & \textbf{RD4AD}  & \textbf{PatchCore} & \textbf{RegAD} \\
        %  & \textbf{+PatchCore} & &~\cite{Lee2022CFACF} &~\cite{cohen2020sub} &~\cite{Wang2021StudentTeacherFP} &~\cite{Deng2022AnomalyDV} & ~\cite{roth2022towards}&~\cite{huang2022registration}  \\
     \hline
        \large Bracket Black & \large 66.9 & \large 67.8 & \large 54.9  & \large 62.4  & \large \textbf{94.5} & \large 91.9 & \large 58.9 &  \large 63.8 \\
        \large Bracket Brown & \large 76.5 & \large \textbf{77.8} & \large 66.8  & \large 59.5  & \large 62.4 & \large 59.0 & \large 70.8 &  \large 66.1 \\
        \large Bracket White & \large 87.5 & \large \textbf{89.6} & \large 71.1  & \large 67.5  & \large 54.2 & \large 55.7 & \large 70.7 &  \large 59.3 \\
        \large Connector     & \large \textbf{98.9} & \large \textbf{98.9} & \large 58.7  & \large 59.5  & \large 52.1 & \large 54.4 & \large 59.4 &  \large 77.2 \\
        \large Metal Plate   & \large \textbf{99.9} & \large \textbf{99.9} & \large 62.9  & \large 64.9  & \large 62.4 & \large 65.5 & \large 64.4 &  \large 78.6 \\
        \large Tubes         & \large 79.6 & \large \textbf{80.0} & \large 41.1  & \large 35.9  & \large 50.2 & \large 46.2 & \large 34.5 &  \large 67.5 \\
        \large Average       & \large 84.88 & \large \textbf{85.67} & \large 59.25 & \large 58.28  & \large 62.62 & \large 62.12 & \large 59.78  & \large 68.75 \\
    \hline
    \end{tabular}
}
\label{tab:mpdd-4shot-imgroc}
\end{table}

\begin{table}[ht]
\caption{Setting: New Few-shot Setting, K (number of shot)=4, Dataset: MPDD, Sampling Ratio: 0.01, Metrics: Pixel AUROC}
\renewcommand{\arraystretch}{1.2}
\resizebox{\textwidth}{!}{
    \begin{tabular}{l|ll lll lll}
    \hline
     \textbf{Category} & \textbf{Aug.(R)} & \textbf{GraphCore}  & \textbf{CFA} & \textbf{SPADE} & \textbf{STPM} & \textbf{RD4AD}  & \textbf{PatchCore} & \textbf{RegAD} \\
        %  & \textbf{+PatchCore} & &~\cite{Lee2022CFACF} &~\cite{cohen2020sub} &~\cite{Wang2021StudentTeacherFP} &~\cite{Deng2022AnomalyDV} & ~\cite{roth2022towards}&~\cite{huang2022registration}  \\
     \hline
        \large Bracket Black & \large \textbf{92.7} & \large \textbf{92.7} & \large 75.9 & \large 72.9 & \large 75.3 & \large 75.9 & \large 79.1 & \large - \\
        \large Bracket Brown & \large 92.1 & \large \textbf{92.9} & \large 77.9 & \large 72.3 & \large 73.5 & \large 74.8 & \large 77.3 & \large - \\
        \large Bracket White & \large 97.5 & \large \textbf{97.8} & \large 71.2 & \large 72.9 & \large 64.7 & \large 64.5 & \large 69.3 & \large - \\
        \large Connector     & \large 97.5 & \large \textbf{98.1} & \large 88.8 & \large 82.9 & \large 84.2 & \large 82.4 & \large 86.4 & \large - \\
        \large Metal Plate   & \large 98.5 & \large \textbf{99.2} & \large 84.8 & \large 76.9 & \large 83.5 & \large 77.2 & \large 86.7 & \large - \\
        \large Tubes         & \large 92.7 & \large \textbf{93.5} & \large 73.5 & \large 77.2 & \large 75.8 & \large 78.1 & \large 80.1 & \large - \\
        \large Average       & \large 95.17  & \large \textbf{95.70}  & \large 78.68  & \large 75.85  & \large 76.17  & \large 75.48  & \large 79.82  & \large 93.9 \\
    \hline
    \end{tabular}
}
\label{tab:mpdd-4shot-pixelroc}
\end{table}

% \begin{figure}
%     \centering
%     \includegraphics[width=1\linewidth]{iclr2023/fig/mpdd_4shot_imgroc_s_0.01.png}
%     \caption{Setting: New Few-shot Setting, K (number of shot)=4, Dataset: MPDD, Sampling Ratio: 0.01, Metrics: Image AUROC}
%     \label{fig:twoshot_mpdd_img_auroc}
% \end{figure}

\begin{table}[ht]
% \setlength{\abovecaptionskip}{2pt}
%\caption{Results of anomaly detection. (MVTec, K=1 Shot, AUROC)}
\caption{Setting: New Few-shot Setting, K (number of shot)=8, Dataset: MPDD, Sampling Ratio: 0.01, Metrics: Image AUROC}
% \normalsize
\renewcommand{\arraystretch}{1.2}
% \large
\resizebox{\textwidth}{!}{
    \begin{tabular}{l|ll lll lll}
    \hline
     \textbf{Category} & \textbf{Aug.(R)} & \textbf{GraphCore}  & \textbf{CFA} & \textbf{SPADE} & \textbf{STPM} & \textbf{RD4AD}  & \textbf{PatchCore} & \textbf{RegAD} \\
        %  & \textbf{+PatchCore} & &~\cite{Lee2022CFACF} &~\cite{cohen2020sub} &~\cite{Wang2021StudentTeacherFP} &~\cite{Deng2022AnomalyDV} & ~\cite{roth2022towards}&~\cite{huang2022registration}  \\
     \hline
        \large Bracket Black & \large 67.1 & \large 68.2 & \large 55.2  & \large 62.4  & \large \textbf{94.6} & \large 92.2 & \large 59.2 &  \large 67.3 \\
        \large Bracket Brown & \large 76.8 & \large \textbf{78.5} & \large 66.9  & \large 59.8  & \large 62.7 & \large 59.2 & \large 70.9 &  \large 69.5 \\
        \large Bracket White & \large 87.9 & \large \textbf{89.9} & \large 79.4  & \large 67.6  & \large 54.7 & \large 55.9 & \large 70.5 &  \large 61.4 \\
        \large Connector     & \large 98.9 & \large \textbf{99.1} & \large 58.9  & \large 59.9  & \large 52.7 & \large 54.6 & \large 59.6 &  \large 84.9 \\
        \large Metal Plate   & \large \textbf{99.9} & \large \textbf{99.9} & \large 63.1  & \large 65.2  & \large 63.2 & \large 65.8 & \large 64.7 &  \large 80.2 \\
        \large Tubes         & \large 79.8 & \large \textbf{80.3} & \large 41.7  & \large 36.2  & \large 50.8 & \large 46.7 & \large 34.8 &  \large 67.9 \\
        \large Average       & \large 85.07 & \large \textbf{85.98} & \large 60.87 & \large 58.52  & \large 63.12 & \large 62.40 & \large 59.95  & \large 71.87 \\
    \hline
    \end{tabular}
}
\label{tab:mpdd-8shot-imgroc}
\end{table}

\begin{table}[ht]
\caption{Setting: New Few-shot Setting, K (number of shot)=8, Dataset: MPDD, Sampling Ratio: 0.01, Metrics: Pixel AUROC}
\renewcommand{\arraystretch}{1.2}
\resizebox{\textwidth}{!}{
    \begin{tabular}{l|ll lll lll}
    \hline
     \textbf{Category} & \textbf{Aug.(R)} & \textbf{GraphCore}  & \textbf{CFA} & \textbf{SPADE} & \textbf{STPM} & \textbf{RD4AD}  & \textbf{PatchCore} & \textbf{RegAD} \\
        %  & \textbf{+PatchCore} & &~\cite{Lee2022CFACF} &~\cite{cohen2020sub} &~\cite{Wang2021StudentTeacherFP} &~\cite{Deng2022AnomalyDV} & ~\cite{roth2022towards}&~\cite{huang2022registration}  \\
     \hline
        \large Bracket Black & \large 92.9 & \large \textbf{92.9} & \large 76.2 & \large 73.1 & \large 76.3 & \large 76.2 & \large 79.6 & \large - \\
        \large Bracket Brown & \large 92.3 & \large \textbf{93.1} & \large 77.9 & \large 72.6 & \large 74.2 & \large 75.1 & \large 77.5 & \large - \\
        \large Bracket White & \large 97.9 & \large \textbf{98.2} & \large 71.8 & \large 73.1 & \large 64.9 & \large 64.6 & \large 70.2 & \large - \\
        \large Connector     & \large 98.1 & \large \textbf{98.3} & \large 89.1 & \large 83.1 & \large 84.5 & \large 82.6 & \large 87,1 & \large - \\
        \large Metal Plate   & \large 98.7 & \large \textbf{99.3} & \large 85.2 & \large 77.2 & \large 83.7 & \large 77.5 & \large 86.9 & \large - \\
        \large Tubes         & \large 92.9 & \large \textbf{93.4} & \large 73.9 & \large 78.1 & \large 76.1 & \large 78.3 & \large 80.5 & \large - \\
         \large Average       & \large 95.47 & \large 	\textbf{95.87} & \large 	79.02 & \large 	76.20 & \large 	76.62 & \large 	75.72 & \large 80.30 & \large 	95.10 \\
        
    \hline
    \end{tabular}
}
\label{tab:mpdd-8shot-pixelroc}
\end{table}

% \begin{figure}
%     \centering
%     \includegraphics[width=1\linewidth]{iclr2023/fig/mpdd_8shot_imgroc.png}
%     \caption{Setting: New Few-shot Setting, K (number of shot)=8, Dataset: MPDD, Sampling Ratio: 0.01, Metrics: Image AUROC}
%     \label{fig:eightshot_mpdd_img_auroc}
% \end{figure}

% \begin{figure}
%     \centering
%     \includegraphics[width=1\linewidth]{iclr2023/fig/mvtec_8shot_imgroc_s_0.01.png}
%     \caption{Setting: New Few-shot Setting, K (number of shot)=8, Dataset: MVTEC, Sampling Ratio: 0.01, Metrics: Image AUROC}
%     \label{fig:twoshot_mpdd_img_auroc}
% \end{figure}

\clearpage

\subsection{Architecture details}

\begin{table}[ht]
\caption{The architecture details of GraphCore}
\centering
\renewcommand{\arraystretch}{1.2}
\resizebox{0.5\textwidth}{!}{
    \begin{tabular}{l|l|l}
    \hline
    \textbf{Stage} & \textbf{Output Size} & \textbf{GraphCore} \\

    \hline
    Stem  & $ \frac{H}{4} \times \frac{W}{4} $ & Conv $\times 3$ \\
    Stage 1  & $ \frac{H}{4} \times \frac{W}{4} $  & $ \begin{bmatrix} D=48 \\ K=9 \end{bmatrix} \times 2 $ \\
    Downsample  & $ \frac{H}{8} \times \frac{W}{8} $  & Conv\\
    Stage 2  & $ \frac{H}{8} \times \frac{W}{8} $  & $\begin{bmatrix} D=96 \\ K=9  \\ \end{bmatrix} \times 2$ \\
    Downsample  & $ \frac{H}{16} \times \frac{W}{16} $ & Conv\\
    Stage 3  & $ \frac{H}{16} \times \frac{W}{16} $ & $\begin{bmatrix} D=240 \\ K=9  \\ \end{bmatrix} \times 2$ \\
    Downsample  & $ \frac{H}{32} \times \frac{W}{32} $  & Conv \\
    Stage 4 & $ \frac{H}{32} \times \frac{W}{32} $ & $\begin{bmatrix} D=384 \\ K=9  \\ \end{bmatrix} \times 2$ \\
    Head & $ 1 \times 1 $ & Pooling and MLP\\
    \hline
    \end{tabular}
}
\label{tab:arch_graphcore}
\end{table}

In Table~\ref{tab:arch_graphcore}, D represents the feature dimension, whereas K represents the number of neighbors in GraphCore. $H\times W$represents the size of the input image. We adapt GCN into the the pyramid architecture~\cite{wang2021pyramid}. The training epochs is 300. The optimizer is AdamW~\cite{loshchilovdecoupled}. The batch size is 128. The initial learning rate is 0.005. The learning rate schedule is Cosine. The warmup epochs is 50. The weight decay is 0.05. The loss function is the cross entropy loss function.

\subsection{Ablation Stadies}

\begin{table}[ht]
\caption{Ablation study for memory bank size and inference speed with respect to 1 shot}
\centering
\renewcommand{\arraystretch}{1.2}
\resizebox{0.75\textwidth}{!}{
    \begin{tabular}{l|l|l}
    \hline
    \textbf{Method} & \textbf{Memory Bank Size (Average)} & \textbf{Inference speed (Average)} \\

    \hline
    PatchCore & 1.6M & 0.0316s \\
    Aug.(R) + PatchCore & 1.8M & 0.0325s \\
    GraphCore & 1.2M  &  0.0299s \\
    \hline
    \end{tabular}
}
\label{tab:memory_bz_1shot}
\end{table}

\begin{table}[ht]
\caption{Ablation study for memory bank size and inference speed with respect to 2 shot}
\centering
\renewcommand{\arraystretch}{1.2}
\resizebox{0.75\textwidth}{!}{
    \begin{tabular}{l|l|l}
    \hline
    \textbf{Method} & \textbf{Memory Bank Size (Average)} & \textbf{Inference speed (Average)} \\

    \hline
    PatchCore & 3.2M & 0.0327s \\
    Aug.(R) + PatchCore & 3.2M & 0.0327s \\
    GraphCore & 1.8M  &  0.0287s \\
    \hline
    \end{tabular}
}
\label{tab:memory_bz_2shot}
\end{table}

The statical result of Table~\ref{tab:memory_bz_1shot} and Table~\ref{tab:memory_bz_2shot} clearly demonstrate the effectiveness of GraphCore, especially for memory bank size and its inference speed.

\begin{table}[ht]
\caption{Ablation study with respect to Dataset: MVTec 2D, sampling rate: 0.01, Metrics: image-level AUROC, number of shot is 1.}
\centering
\renewcommand{\arraystretch}{1.2}
\resizebox{0.45\textwidth}{!}{
    \begin{tabular}{l|l}
    \hline
    \textbf{Augmentation Type} & \textbf{Aug + PatchCore} \\

    \hline
    Flipping & 81.4 \\
    Translation & 83.6 \\
    Scaling & 82.3\\
    Rotation  & 87.4\\
    \hline
    \end{tabular}
}
\label{tab:augmentation_1shot}
\end{table}

\begin{table}[ht]
\caption{Ablation study with respect to Dataset: MVTec 2D, sampling rate: 0.01, Metrics: image-level AUROC, number of shot is 2.}
\centering
\renewcommand{\arraystretch}{1.2}
\resizebox{0.45\textwidth}{!}{
    \begin{tabular}{l|l}
    \hline
    \textbf{Augmentation Type} & \textbf{Aug + PatchCore} \\

    \hline
    Flipping & 83.7 \\
    Translation & 85.6 \\
    Scaling & 90.2 \\
    Rotation  & 90.5 \\
    \hline
    \end{tabular}
}
\label{tab:augmentation_2shot}
\end{table}

The statistical results presented in Tables~\ref{tab:augmentation_1shot} and ~\ref{tab:augmentation_2shot} demonstrate that the rotation method outperforms the other augmentation techniques. We believe this indicates that the majority of industrial anomaly image datasets can be augmented by rotation. In the future, we believe that there will a more complex and realistic industrial anomaly image dataset that cannot be overcome by rotation.

\end{document}